%% file: beauty.tex
\documentclass[10pt,twocolumn,letterpaper]{article}

\usepackage{iccv}
\usepackage{times}
\usepackage{epsfig}
\usepackage{graphicx}
\usepackage{amsmath}
\usepackage{amssymb}
\usepackage{soul}

\usepackage{enumitem}
\usepackage{booktabs}
\usepackage{array}
\usepackage{caption,subcaption}
\usepackage{lipsum}
\usepackage[pagebackref=true,breaklinks=true,letterpaper=true,colorlinks,bookmarks=false]{hyperref}

\DeclareGraphicsExtensions{.pdf,.jpg,.png}
\graphicspath{{figs/}}

\newcommand{\figref}[1]{Figure~\ref{fig:#1}}
\newcommand{\tblref}[1]{Table~\ref{tbl:#1}}

\iccvfinalcopy 

\def\iccvPaperID{2954} 

\ificcvfinal\fi

\begin{document}

\title{Understanding and Mapping Natural Beauty}

\author{
  \begin{minipage}{\linewidth}
    \centering
    \begin{minipage}{1.5in}
      \centering
      Scott Workman$^1$\\
      {\tt\small scott@cs.uky.edu}
    \end{minipage}
    \begin{minipage}{1.5in}
      \centering
      Richard Souvenir$^2$\\
      {\tt\small souvenir@temple.edu}
    \end{minipage}
    \begin{minipage}{1.5in}
      \centering
      Nathan Jacobs$^1$\\
      {\tt\small jacobs@cs.uky.edu}
    \end{minipage}
    \\[.2cm]
    \begin{minipage}{1.7in}
      \centering
      $^1$University of Kentucky \\
    \end{minipage}
    \begin{minipage}{1.7in}
      \centering
      $^2$Temple University \\
    \end{minipage}
  \end{minipage}
}

\maketitle

\begin{abstract}
  While natural beauty is often considered a subjective property of
  images, in this paper, we take an objective approach and provide
  methods for quantifying and predicting the scenicness of an image.
  Using a dataset containing hundreds of thousands of outdoor images
  captured throughout Great Britain with crowdsourced ratings of
  natural beauty, we propose an approach to predict scenicness which
  explicitly accounts for the variance of human ratings. We
  demonstrate that quantitative measures of scenicness can benefit
  semantic image understanding, content-aware image processing, and a
  novel application of cross-view mapping, where the sparsity of
  ground-level images can be addressed by incorporating unlabeled
  overhead images in the training and prediction steps. For each
  application, our methods for scenicness prediction result in
  quantitative and qualitative improvements over baseline approaches.
\end{abstract}

\section{Introduction}

Recent advances in learning with large-scale image collections 
have led to methods that go beyond identifying objects and 
their interactions toward quantifying seemingly
subjective high-level properties of the scene. For example, Isola et
al.~\cite{isola2011makes} explore image memorability, finding that 
memorability is a stable property of images that can be predicted 
based on the image attributes and features. Other similar high-level 
image properties include photographic style~\cite{su2011scenic},
virality~\cite{deza2015understanding},
specificity~\cite{jas2015image}, and
humor~\cite{chandrasekaran2016we}. Quantifying such properties
facilitates new applications in image understanding.

\begin{figure}
  \centering
  \includegraphics[width=.49\linewidth]{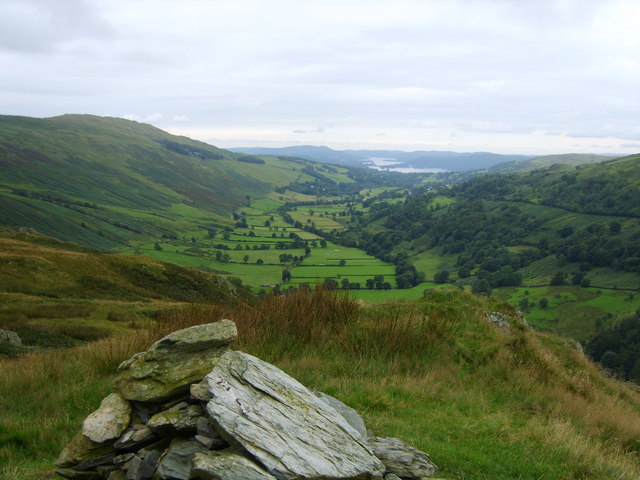}
  \includegraphics[width=.49\linewidth]{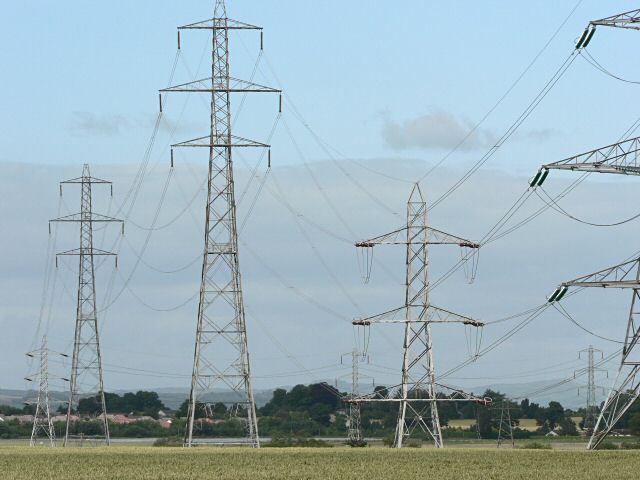}
  \includegraphics[width=.49\linewidth]{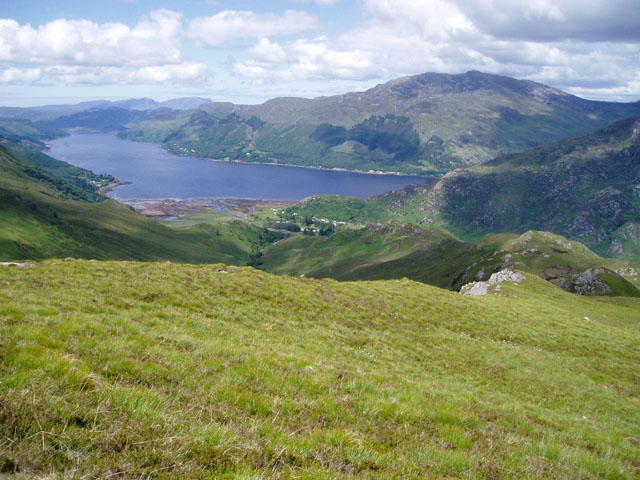}
  \includegraphics[width=.49\linewidth]{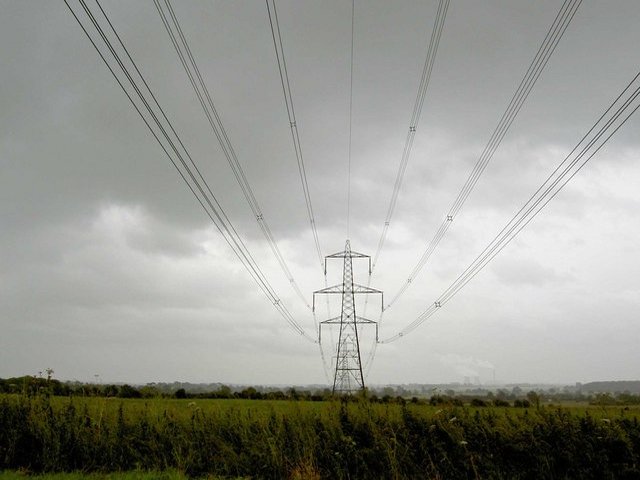}
  
  \caption{Most observers agree that images of mountains are more scenic than
	power lines. Our work
  seeks to automatically quantify ``scenicness''
	and demonstrate applications in image understanding and mapping.}
  \label{fig:cartoon}
\end{figure}

In this paper we consider ``scenicness'', or the natural beauty
of outdoor scenes. 
Despite the popularity of the saying ``beauty lies in the eye
of the beholder,'' research shows that beauty is not purely 
subjective~\cite{langlois2000maxims}. For example, consider the 
images in \figref{cartoon}; mountainous 
landscapes captured from an
elevated position are consistently rated as more beautiful by humans than
images of power transmission towers.

Understanding the perception of landscapes has been an active
research area (see~\cite{zube1982landscape} for a comprehensive review) 
with real-world importance.  For example,
McGranahan~\cite{mcgranahan1999natural} derives a natural amenities
index and shows that rural population change is strongly related to
the attractiveness of a place to live, as well as an area's popularity
for retirement or recreation. Seresinhe et
al.~\cite{seresinhe2015quantifying} show that inhabitants of more
beautiful environments report better overall health. Runge et al.~\cite{runge2016no}
characterize locations by their visual attributes and describe a
system for scenic route planning. Lu et al.~\cite{lu2010photo2trip}
recover cues from millions of geotagged photos to
suggest customized travel routes.

\begin{figure*}
  \centering

  \newlength{\myheight}
  \setlength{\myheight}{.32\columnwidth}
  
  \begin{subfigure}{.49\linewidth}
    \centering
    \includegraphics[width=\myheight]{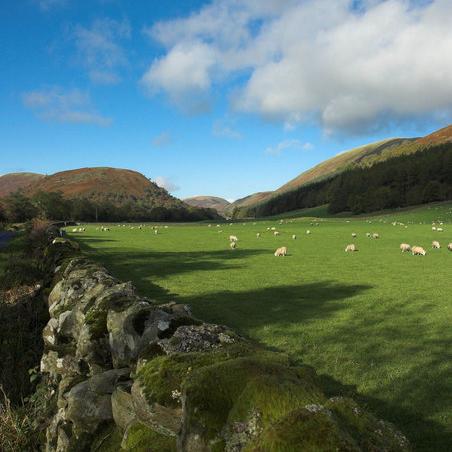}
    \includegraphics[width=\myheight]{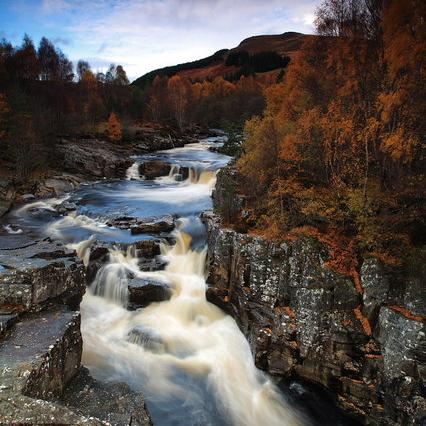}
    \includegraphics[width=\myheight]{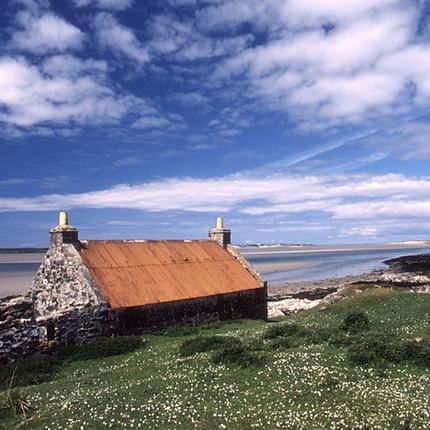}
    \includegraphics[width=\myheight]{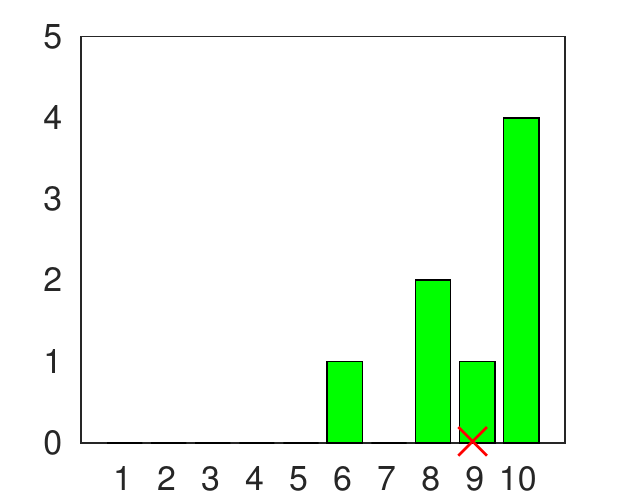}
    \includegraphics[width=\myheight]{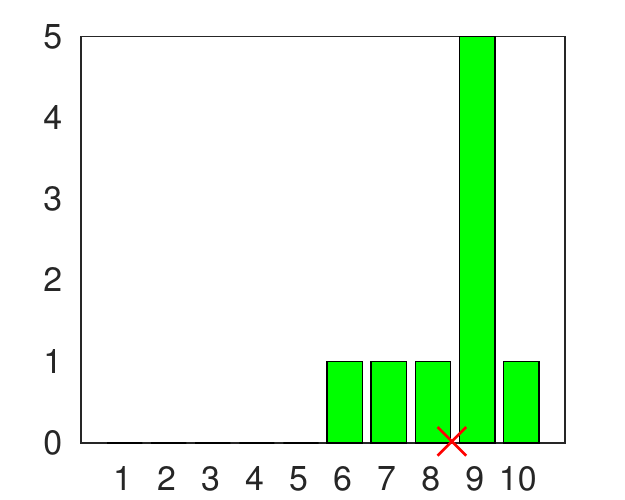}
    \includegraphics[width=\myheight]{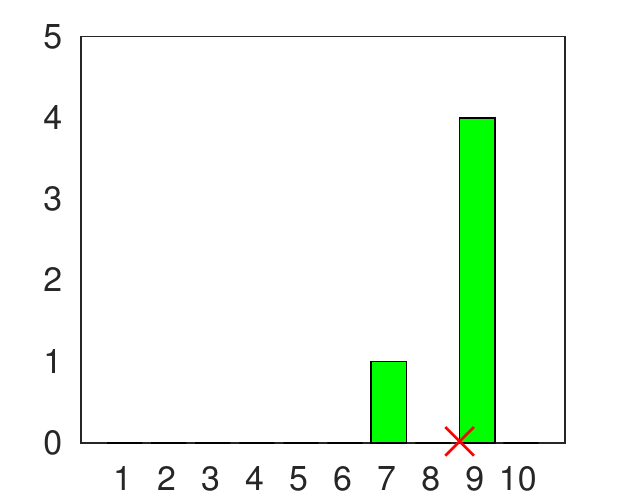}
    \caption{Scenic}
  \end{subfigure}
  \begin{subfigure}{.49\linewidth}
    \centering
    \includegraphics[width=\myheight]{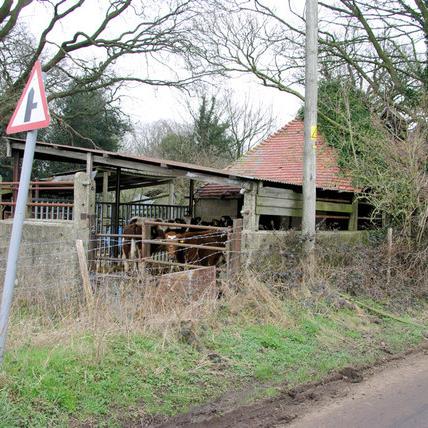}
    \includegraphics[width=\myheight]{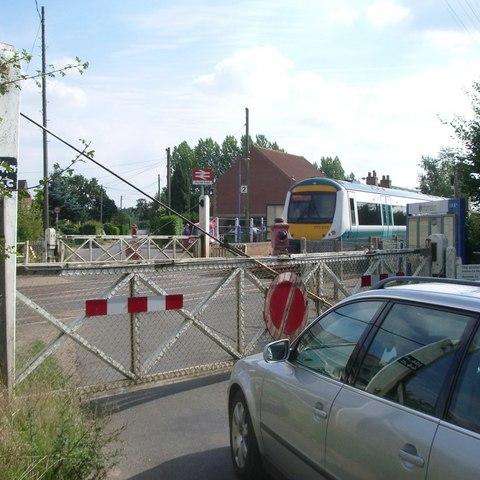}
    \includegraphics[width=\myheight]{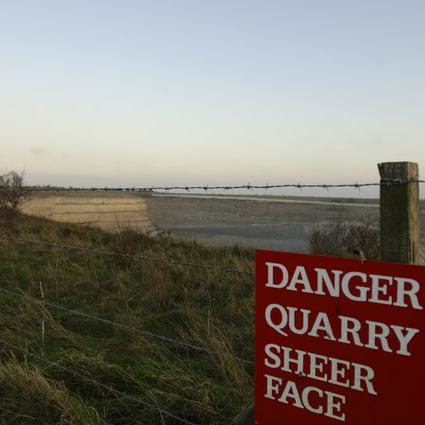}
    \includegraphics[width=\myheight]{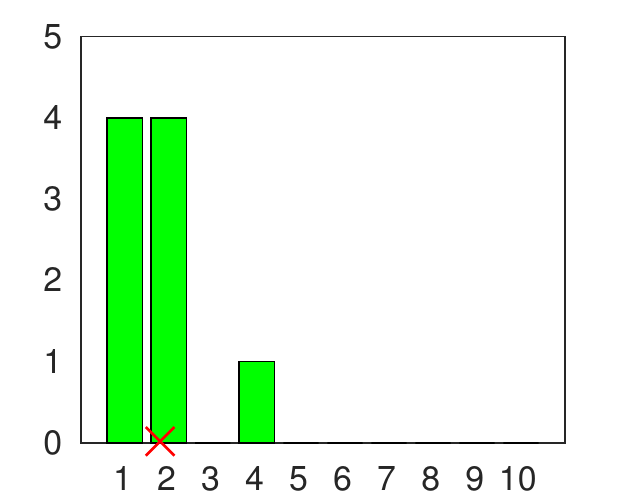}
    \includegraphics[width=\myheight]{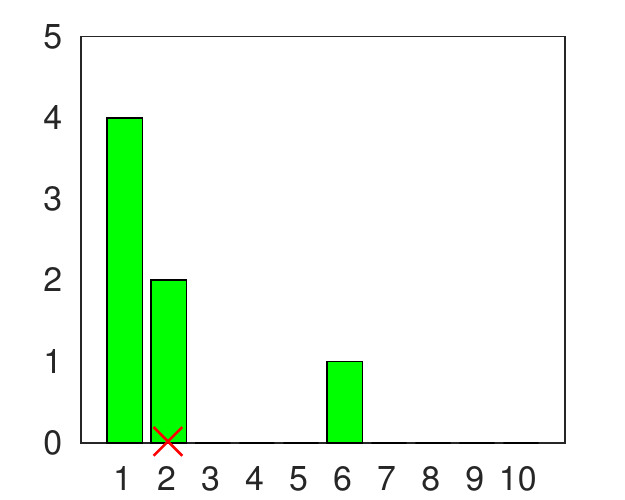}
    \includegraphics[width=\myheight]{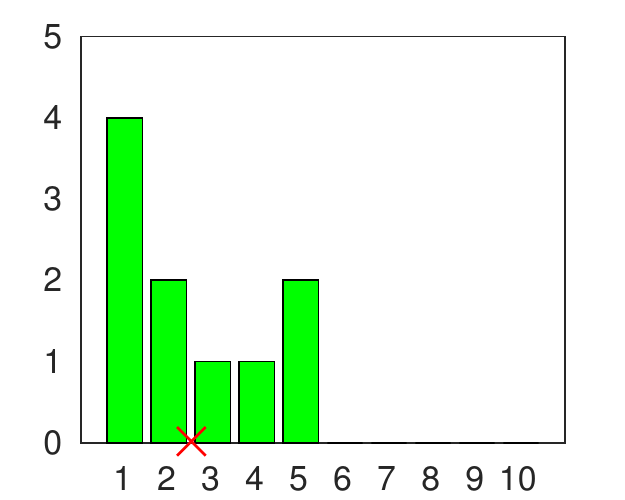}
    \caption{Non-Scenic}
  \end{subfigure}

  \caption{Example images (and human-provided scenicness ratings) from the ScenicOrNot (SoN) dataset: (a) ``scenic''
  images (average rating above 7.0) and (b) ``non-scenic'' images
  (average rating below 3.0).}%
\label{fig:SoNimages}%
\end{figure*}

Recently, a number of algorithms have been developed to automatically
interpret high-level properties of images. Laffont et
al.~\cite{laffont14transient} introduce a set of transient
scene attributes and train regressors for estimating them in novel
images. Lorenzo et al.~\cite{lorenzo2015safety} use a convolutional
neural network to estimate urban perception from a single image. Deza
and Parikh~\cite{deza2015understanding} study the phenomenon of image
virality. Similarly, a significant amount of work has sought to
understand the relationship between images and their
aesthetics~\cite{ke2006design,luo2008photo,yeh2010personalized}.
Karayev et al.~\cite{karayev2014style} recognize photographic style.
Su et al.~\cite{su2011scenic} propose a method for scenic photo
quality assessment using hand-engineered features. Developed
independently from our work, Seresinhe et
al.~\cite{seresinhe2017quantifying,seresinhe2017using} explore models
for quantifying scenicness. Lu et al.~\cite{lu2014rapid} apply deep
learning to rate images as high or low aesthetic quality.

In this paper, we start with a large-scale dataset containing hundreds
of thousands of images, individually rated by humans, to
quantify and predict image scenicness. 
Our main contributions are: 
\begin{itemize}[itemsep=0pt]
  \item an analysis of outdoor images to identify
    semantic concepts correlated with scenicness;
  \item a method for estimating the scenicness of an image which
    accounts for variance in the ratings and human perception of
    scenicness; 
\item a new dataset of ground-level and overhead images with
    crowdsourced scenicness scores; and  
 \item a novel cross-view mapping approach, which incorporates both
    ground-level and overhead imagery to address the spatial sparsity
		of ground-level images, to provide country-scale predictions of scenicness.
\end{itemize}

\section{Exploring Image Scenicness}

Our work builds on a publicly-available crowd-sourced database
collected as part of an online game,
ScenicOrNot,\footnote{ScenicOrNot~(\url{http://scenicornot.datasciencelab.co.uk/})
is built on top of Geograph~(\url{http://www.geograph.org.uk/}), an
online community and photo-sharing website.} which contains images
captured throughout Great Britain. As part of the game, users are
presented a series of images from around the island of Great Britain
and invited to rate them according to their scenicness, or natural
beauty, on a scale from 1-10. From a user standpoint, in addition to
being exposed to the diverse environments of England, Scotland and
Wales, the purpose of the game is to compare aesthetic judgments
against those of other users.

We apply our work to a database of 185,548 images and associated
natural beauty rating histograms. Each image in the dataset was rated at least
five times.  We refer to this set of images as the ScenicOrNot (SoN)
dataset. In addition to retaining the rating distribution and average
rating, we partition the set of images into ``scenic'' (average rating
above 7.0) and ``non-scenic'' (average rating below 3.0) subgroups.
\figref{SoNimages} shows sample images from the dataset. In the
remainder of this section, we explore image properties that may be associated with
scenicness, including: text annotations, color
statistics, and semantic image attributes.

\begin{figure}
  \centering
  \includegraphics[clip,trim=1.5cm 1.5cm 1.5cm .5cm,width=\linewidth]{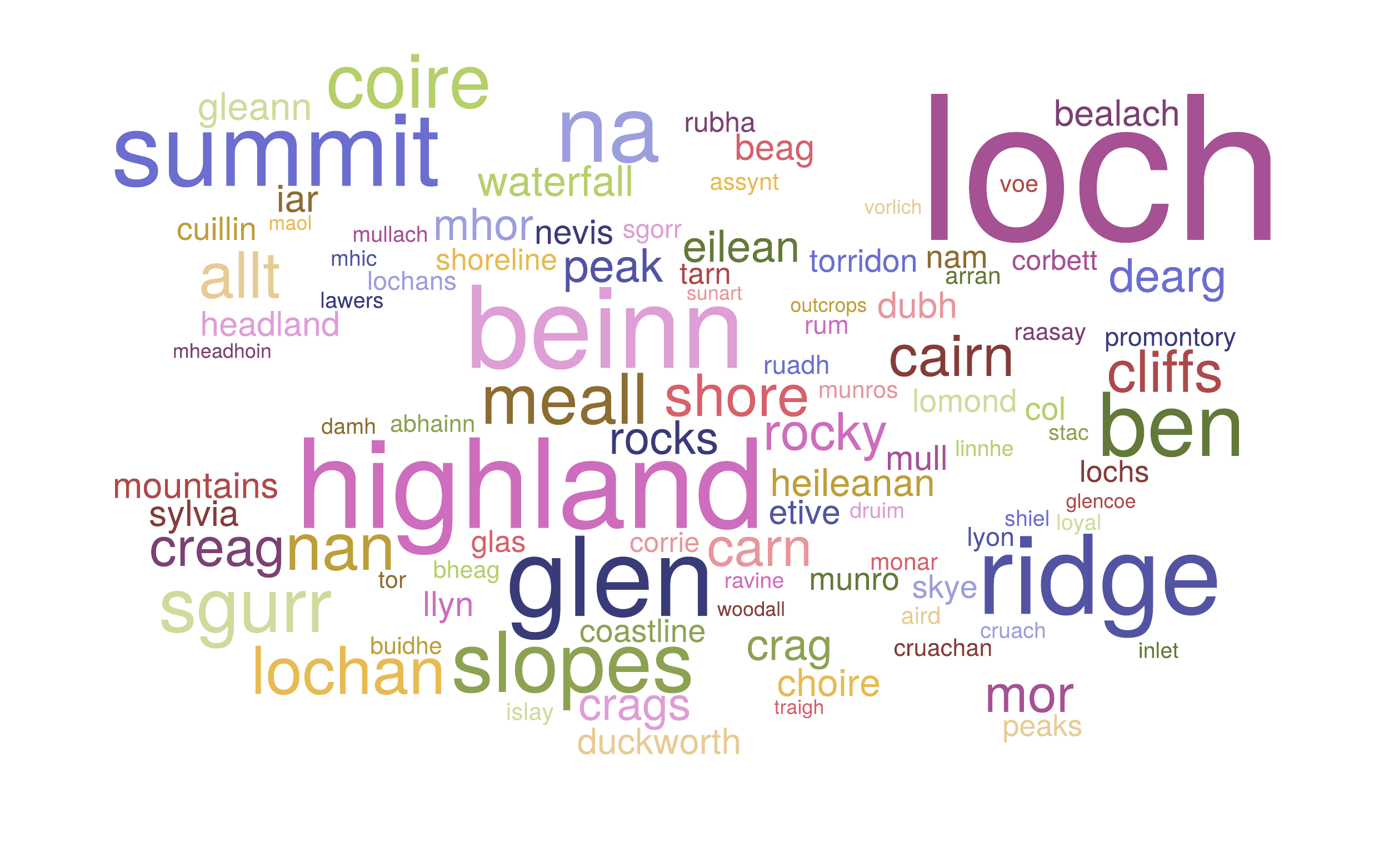}
  \caption{The word cloud depicts the relative frequency of title and caption terms
	found in scenic images from the SoN dataset.}
  \label{fig:wordcloud}
\end{figure}

\subsection{Image Captions}

Like most images hosted on image sharing sites, the SoN images have
associated metadata, including a title and caption. For example, the
image in \figref{cartoon} (top, left) is titled {\em From Troutbeck
Tongue} and has the following caption: ``Looking over the cairn down
Trout Beck. Windermere and the sea in the distance". For all of the images
in the SoN dataset, we analyzed the title and captions to explore whether these
associated text annotations are correlated with scenicness.

Using the scenic and non-scenic subsets, we compute the relative term
frequency for each of the extracted words.  \figref{wordcloud} shows a
word cloud of the most frequent 100 extracted terms from scenic
images, where the size of the word represents the relative frequency.
While some of the terms (\eg, ``ridge'', ``cliffs'', ``summit'') may
universally correlate with scenicness, other terms, such as ``loch'',
``na'', and ``beinn'' reflect the fact the data originates from Great
Britain. Conversely, example terms that are negatively correlated with
scenicness include ``road'', ``lane'', ``house'', and ``railway''.

\subsection{Color Distributions}

\begin{figure}
  \centering
  \includegraphics[clip,trim=1.2cm .2cm .85cm .5cm,width=.8\linewidth]{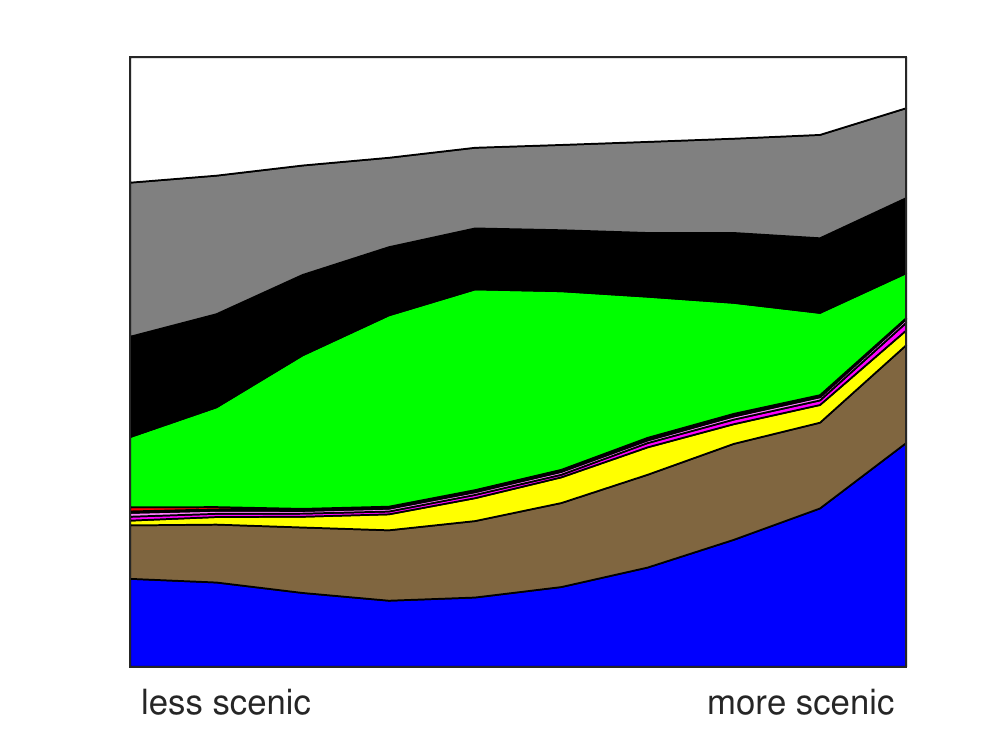}
  \caption{Distribution of color with respect to the average scenicness rating of
	the SoN image set.}
  \label{fig:colorhistogram}
\end{figure}

The images in \figref{SoNimages} and terms in \figref{wordcloud}
suggest that images with blue skies, green fields, water, and other
natural features tend to be rated as more scenic. For this analysis,
we computed the distribution of quantized color values, using the
approach of Van De Weijer et al.~\cite{van2009learning}, as a function
of the average scenicness rating of the SoN image set.
\figref{colorhistogram} shows the distribution, where we see blue
overrepresented in scenic images and, conversely, black and gray
overrepresented in non-scenic images.

\subsection{Scene Semantics} 

\begin{figure}
  \centering
  \includegraphics[clip,trim=2cm .4cm .6cm .5cm,width=\linewidth]{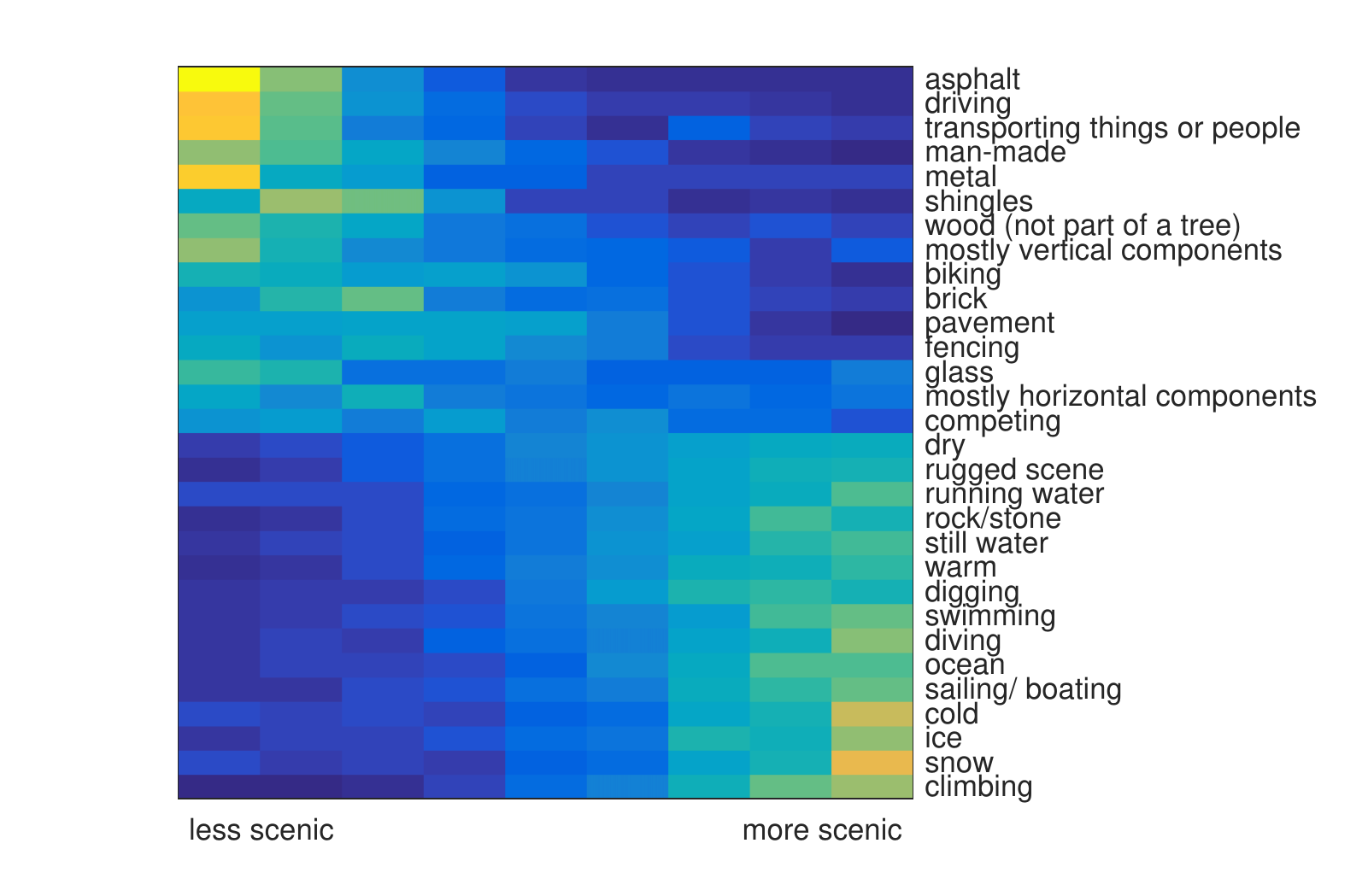}
  \caption{Distribution of the frequency of SUN
  attributes~\cite{patterson2012sun} in ``scenic'' versus ``not
  scenic'' images. Warm colors indicate higher frequency.}
  \label{fig:attr_occurrence}
\end{figure}

For each image, we compute SUN attributes~\cite{patterson2012sun}, a
set of 102 discriminative scene attributes spanning several types (\eg,
function, materials).  \figref{attr_occurrence} shows an
occurrence matrix for a subset of attributes correlated with image scenicness. 
Attributes such as ``asphalt'',
``man-made'', and ``transporting things or people'' occur often in
less scenic images, suggesting that urban environments are more
typical of images with low scenicness. In contrast, attributes such as
``ocean'', ``climbing'', and ``sailing/boating'' occur more often in the most
scenic images.

\begin{figure}
  \centering
  \includegraphics[clip,trim=1.1cm .2cm .6cm .4cm,width=\linewidth]{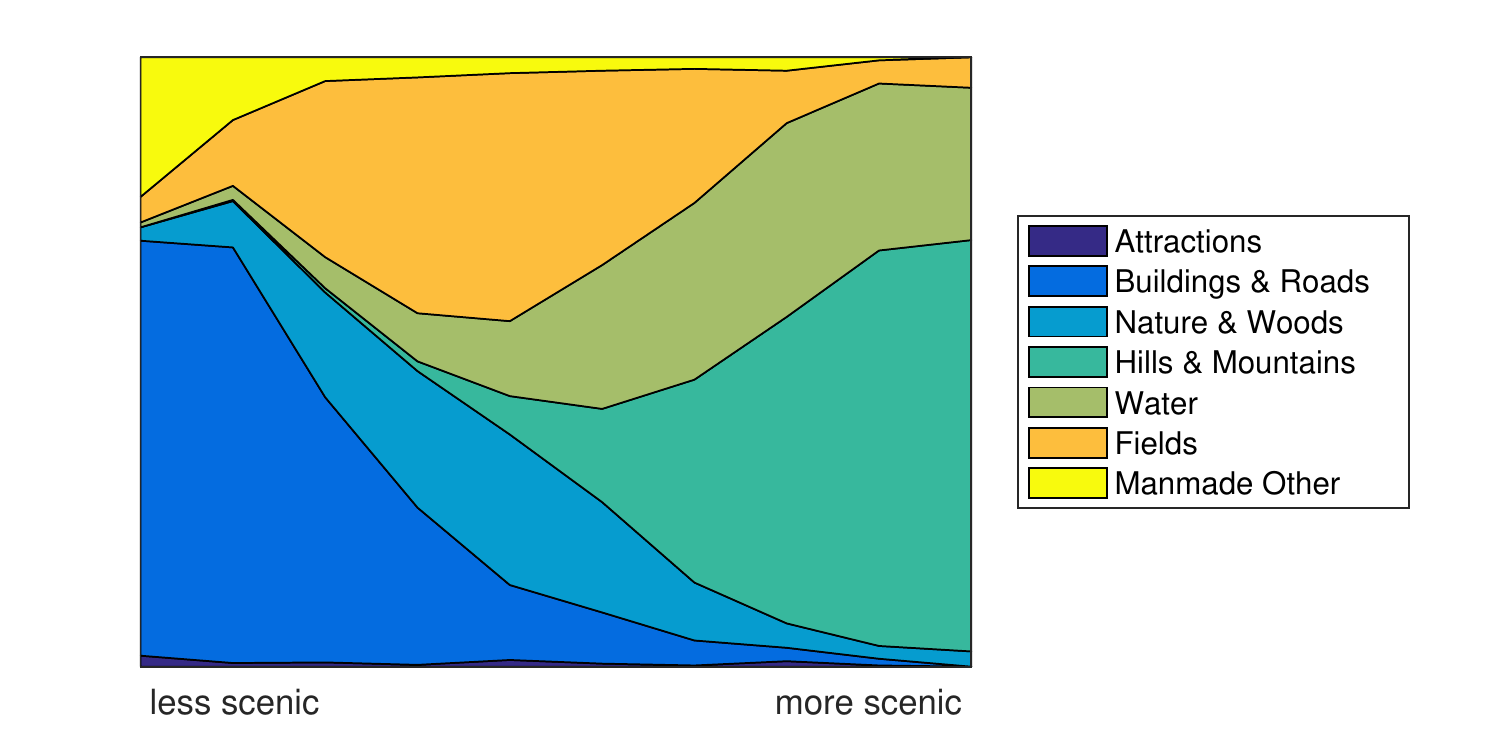}
  \caption{Distribution of high-level categories for the images in the
  SoN dataset.}
  \label{fig:places_vs_beauty}
\end{figure}

Similarly, we compared scenicness to the scene categorizations
generated by the Places~\cite{zhou2014places} convolutional neural
network.  Of the 205 Places scene
classes (\eg, ``airplane cabin'', ``hotel room'', ``shed''), 135 describe outdoor
categories.  We aggregate the outdoor classes into seven higher-level
scene categories (similar to Runge et al.~\cite{runge2016no}), such as
``buildings and roads'', ``nature and woods'', and ``hills and
mountains''.  Each image is classified using Places into one of these
high-level categories.  \figref{places_vs_beauty} shows the frequency
of each category as a function of the average user-provided
rating of SoN images. The trend follows previously observed
patterns; on the whole, images containing natural features, such as
hills, mountains, and water, are rated as more scenic than images
containing buildings, roads, and other man-made constructs.

\subsection{Summary}

This analysis shows that scenicness is related to both low-level image
characteristics, such as color, and semantic properties, such as
extracted attributes and scene categories.  
This suggests that it is possible to estimate scenicness
from images.  In the following section, we
propose a method for directly estimating image scenicness from raw
pixel values.

\section{Predicting Image Scenicness}
\label{sec:method}

We use a deep convolutional neural network (CNN) to address the task
of automatically estimating the scenicness of an image. Following
other approaches (\eg, \cite{planet,workman2016horizon}), we
partition the output space and treat this prediction as 
a discrete labeling task where the output layer corresponds to the
integer ratings (\ie, $1,2,\ldots,10$) of scenicness. We
represent our CNN as a function, $G(I;\Theta_g)$, where $I$ is an
image and the output is a probability distribution over the 10
scenicness levels. We consider multiple loss
functions during training to best capture the distribution in human ratings of scenicness for a given image.

The baseline approach follows recent work (\eg,~\cite{xie2011im2map}), which trains
a model to predict a
single value. For this variant, each image is associated with the 
label corresponding to the mean human rating, rounded to the nearest integer value, $\bar{r}$. Training involves minimizing 
the typical cross-entropy loss:
\begin{align} 
  E = -\frac{1}{N}\sum_{n=1}^{N} \log(G(I_n;\Theta_g)(\bar{r}_n)),
  \label{eq:logloss} 
\end{align} 
where $N$ is the number of training examples.

The baseline approach assumes a single underlying
value for scenicness. However, as shown in \figref{SoNimages}, for many images, there may be 
high variability 
in the ratings. In these cases, the mean scenicness 
may not serve as a representative value. So, instead of directly
predicting the mean scenicness, we train the model to 
predict the human rating distribution for a particular image.
For this variant, we treat the normalized human ratings as a target distribution
and train the model to predict this distribution directly, by
minimizing the cross-entropy loss:  
\begin{align} 
  E = -\frac{1}{N}\sum_{n=1}^{N} \sum_{r=1}^{10} p_{nr} \log(G(I_n;\Theta_g)(r)),
	\label{eq:distloss} 
\end{align}
where $p_{nr}$ is the proportion of $r$ ratings for image $n$.

However, the previous formulation assumes a large number of ratings so
that $p_n$ approaches the true distribution.
In our case, this assumption does not hold. 
As an alternative to 
predicting the mean scenicness or the empirical scenicness distribution, we 
model the set of ratings for an image as a sample from a multinomial
distribution. Each 
training example is associated
with a set of (potentially noisy) labels
$\{(I_1,\{v_{1i}\}),\ldots,(I_N,\{v_{Ni}\})\}$, where $\{v_{ji}\}$ is
the set of ratings for image $I_j$. 
This results in the following loss:
\begin{align} 
  E = -\frac{1}{N}\sum_{n=1}^{N} \sum_{i=1}^{V_n} p_{ni} \log(G(I_n;\Theta_g)(v_{ni})),
	\label{eq:ourloss}
\end{align}
where $V_n$ is the total number of ratings for image $n$.

\subsection{Comparison with Human Ratings}
\label{sec:evaluation}

We evaluate our scenicness predictions using the SoN dataset. We
reserved 1,413 images that have at least ten ratings as test cases
for evaluation, with the remaining data used for training and
validation. For predicting scenicness, we modify the GoogleNet
architecture~\cite{szegedy2015going} with weights initialized from the
{\em Places} network~\cite{zhou2014places}. We selected this CNN
because it had been trained for the related task of outdoor scene
classification; however, our methods could be applied to other
related architectures or trained from scratch with sufficient data. Our
implementation uses the Caffe~\cite{jia2014caffe} deep learning
toolbox. For training, we randomly initialize the last layer weights
and optimize parameters using stochastic gradient descent with a base
learning rate of $10^{-4}$ and a mini-batch size of 40 images. Roughly
10\% of the training data is reserved for validation. All trained
models, including example code, are available at our project
website.\footnote{\label{projectsite}\url{http://cs.uky.edu/~scott/research/scenicness/}}

We refer to the three models as: (1) {\sc Average}, the baseline approach
that predicts the mean scenicness (Equation~\ref{eq:logloss}); (2) {\sc Distribution}, the model that
minimizes cross-entropy loss to the normalized distribution of human ratings (Equation~\ref{eq:distloss}); and
(3) {\sc Multinomial}, which maximizes the multinomial
log-likelihood (Equation~\ref{eq:ourloss}). We compare performance on two tasks: (1) predicting
the average human rating and (2) predicting the distribution
of ratings for a given image.

The output of each network is a posterior probability for each integer rating
for a given input image. To evaluate the average user predictions, we consider
the order of the predictions, ranked by posterior probability and use 
the information retrieval metric, 
\emph{Normalized Discounted Cumulative Gain (nDCG)}, which penalizes ``out of order'' 
posterior probabilities, given the ground-truth rating. The second column of Table~\ref{tab:humanRatings} shows the nDCG scores for each of the three models. Overall, the models trained using 
different loss functions performed similarly well under this evaluation metric.

\begin{table}%
  \centering
  \caption{Quantitative results comparing models with different loss
  functions. For each metric, higher is better.}
  \begin{tabular}{@{}lcc@{}}
    \hline
    Loss & \multicolumn{2}{c}{Metric} \\
         \cline{2-3}
         & nDCG & K-S \\
    \hline
    {\sc Average} & .9780 & 14.8\% \\
    {\sc Distribution} & .9678 & 50.0\% \\
    {\sc Multinomial} & .9745 & 58.4\% \\
    \hline
  \end{tabular}
  \label{tab:humanRatings}
\end{table}

For the task of predicting the distribution of ratings for a given image, the
performance of the models diverged. We take
a hypothesis testing approach and consider whether or not the set of human
ratings could be drawn from the distribution represented by the output probabilities
of the CNN. For this, we applied the one-sample \emph{Kolmogorov-Smirnov (K-S)} test with a
non-parametric distribution and computed the proportion of testing images
for which the human ratings come from the posterior distribution at the 5\% significance level.
The last column of Table~\ref{tab:humanRatings} shows the percentage of testing images that matched the predicted distribution.
The models trained using distribution of ratings, {\sc Distribution} and {\sc Multinomial}, significantly outperform the model trained on average 
rating, with {\sc Multinomial} showing the best performance. 

\begin{figure}

  \centering

  \newcolumntype{M}{>{\centering\arraybackslash}m{\dimexpr.48\linewidth-2\tabcolsep}}
  \begin{tabular}{MM}
    \includegraphics[width=.96\linewidth]{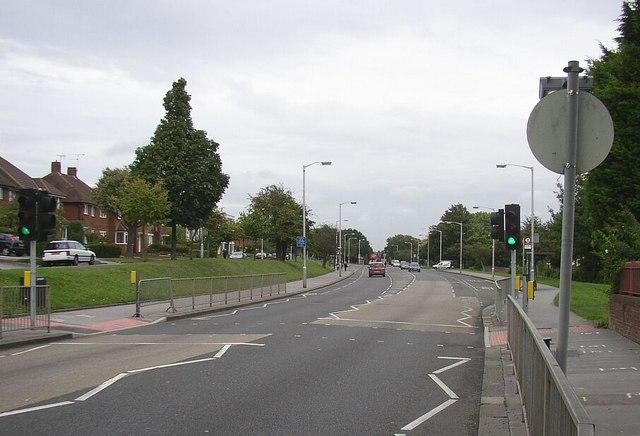} &
    \includegraphics[width=.96\linewidth]{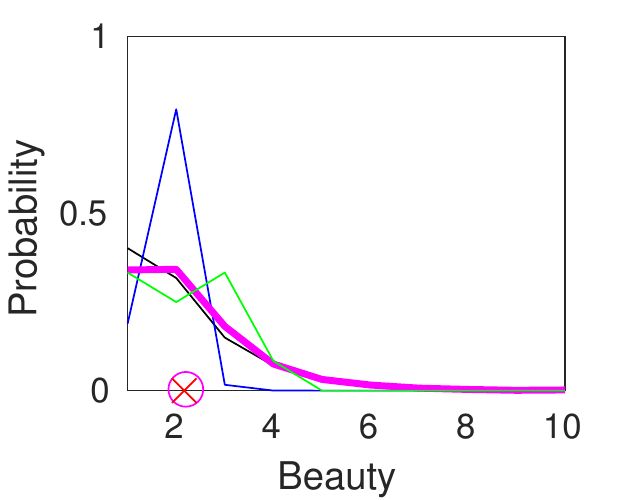} \\
    \includegraphics[width=.96\linewidth]{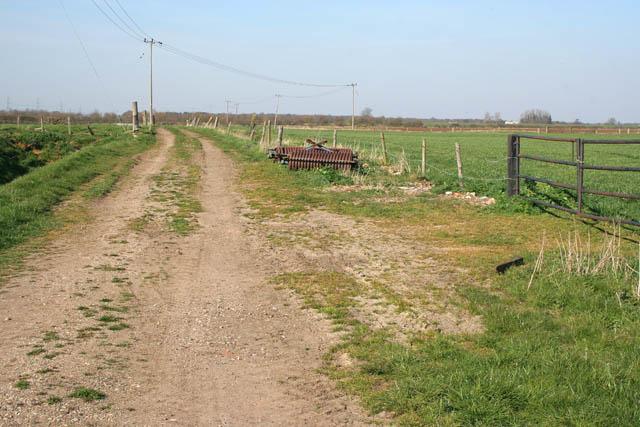} &
    \includegraphics[width=.96\linewidth]{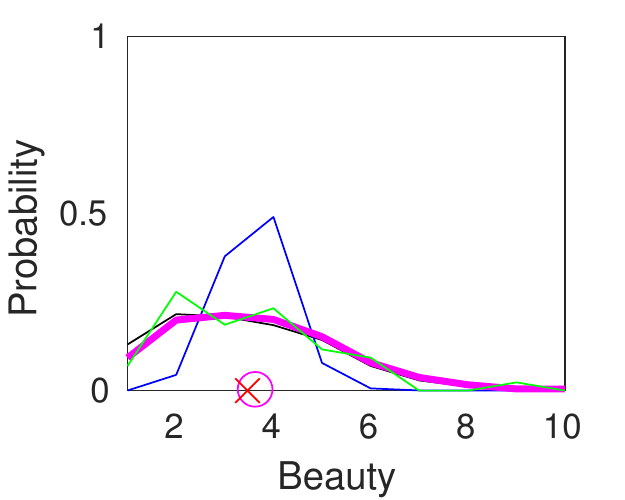} \\
    \includegraphics[width=.96\linewidth]{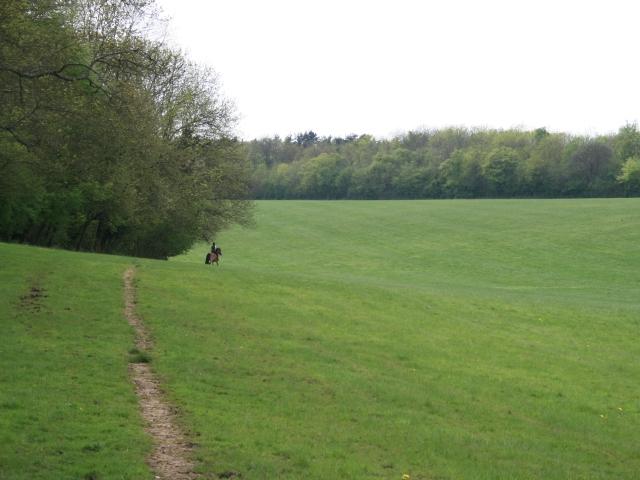} &
    \includegraphics[width=.96\linewidth]{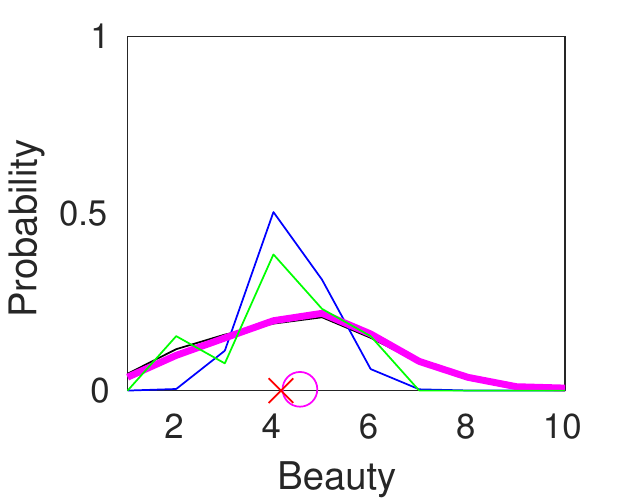} \\
    \includegraphics[width=.96\linewidth]{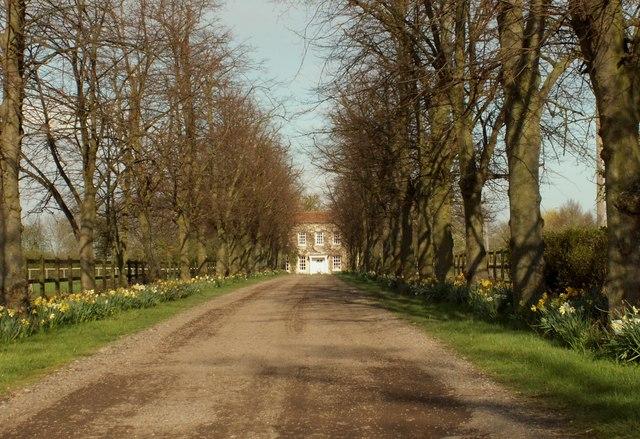} &
    \includegraphics[width=.96\linewidth]{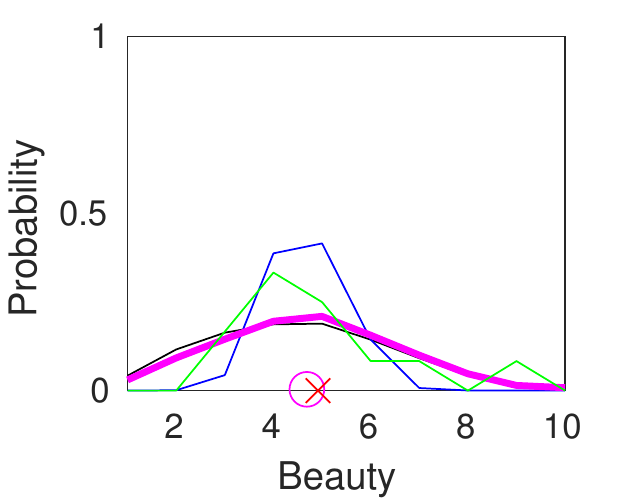} \\
    \includegraphics[width=.96\linewidth]{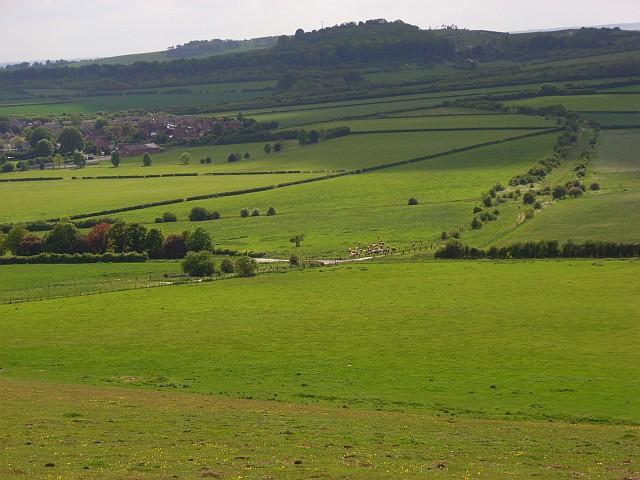} &
    \includegraphics[width=.96\linewidth]{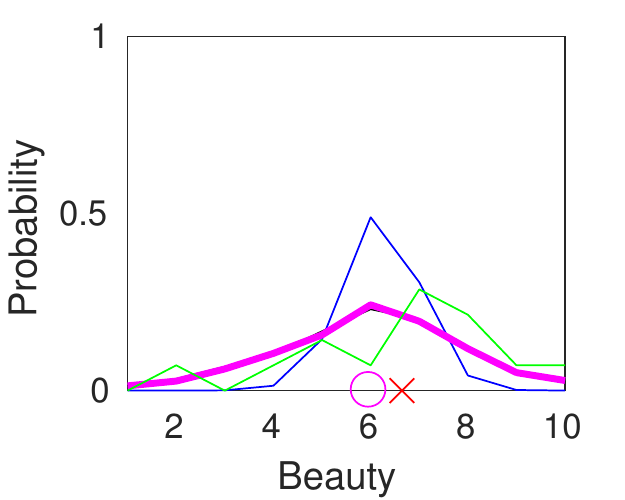} \\
    \includegraphics[width=.96\linewidth]{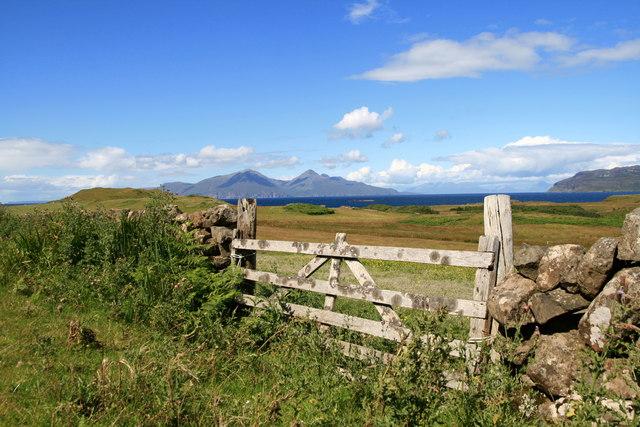} &
    \includegraphics[width=.96\linewidth]{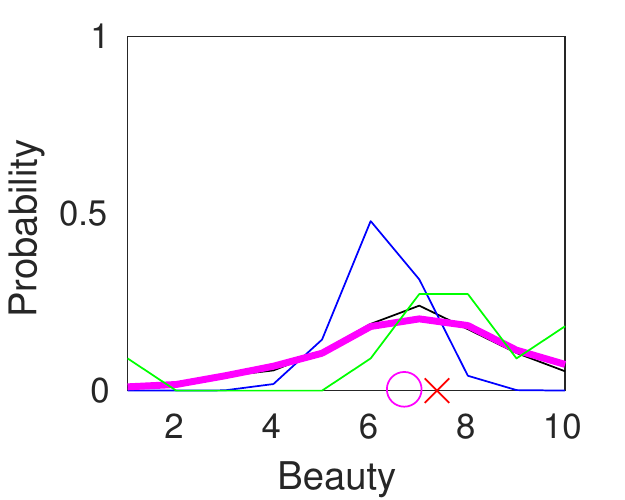} \\
    \includegraphics[width=.96\linewidth]{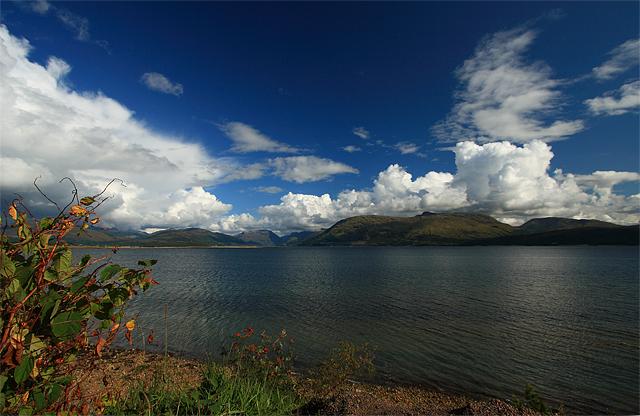} &
    \includegraphics[width=.96\linewidth]{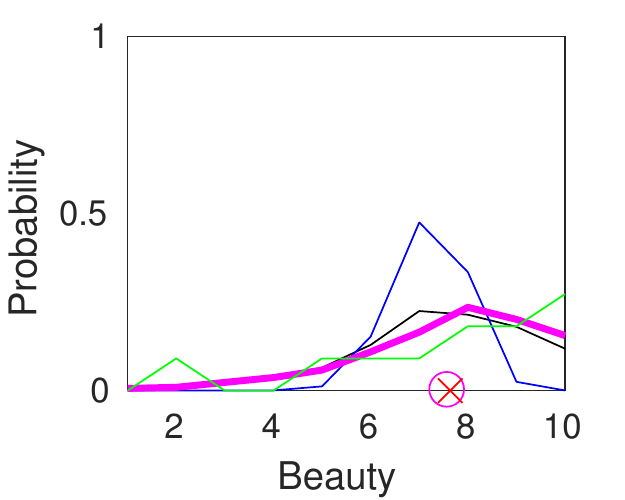}
  \end{tabular}

  \caption{Example images alongside the distribution of human ratings
  (green), and the outputs of {\sc Average} (blue), {\sc Distribution} 
	(black), and {\sc Multinomial} (magenta). The red~$\times$ corresponds to
  the mean rating and the
  magenta~$\circ$ the weighted average of the {\sc Multinomial} prediction.
}

  \label{fig:uncertainty}
\end{figure}

\figref{uncertainty} visualizes these results qualitatively. Several
example images are shown alongside the distribution of human ratings 
(green) and predictions from the three models. In general, the results
follow the quantitative analysis. The {\sc Multinomial} method better
captures human uncertainty as compared to the other methods. For example, in
\figref{uncertainty} (row 1), the baseline approach, {\sc Average}, provides a 
much higher posterior probability for a rating of 2 than 
the distribution of humans ratings. Comparatively, {\sc Multinomial} is more
consistent with human
ratings and closer to the average user predictions. For the remaining
experiments, the {\sc Multinomial} model is used unless otherwise specified.

\subsection{Receptive Fields of Natural Beauty} 
 
\begin{figure}
  \centering
  \begin{subfigure}{1\linewidth}
    \centering
    \includegraphics[width=.32\linewidth]{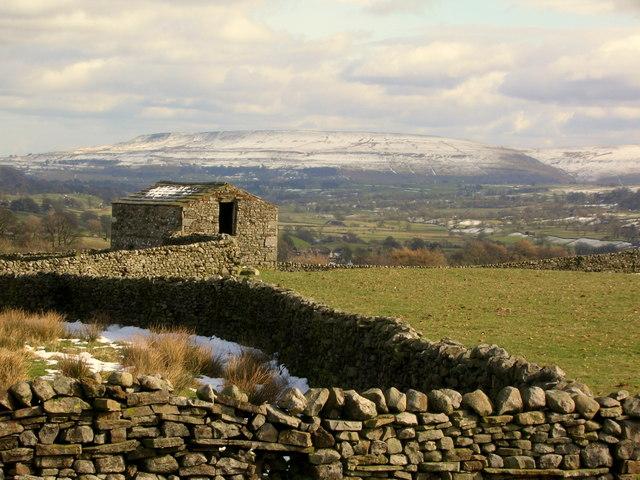}
    \includegraphics[width=.32\linewidth]{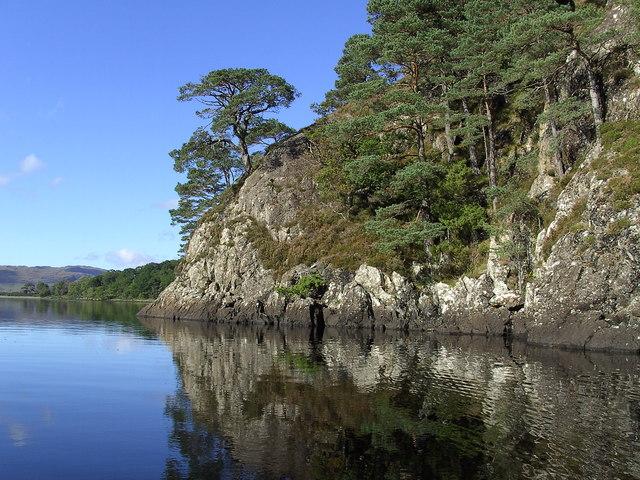}
    \includegraphics[width=.32\linewidth]{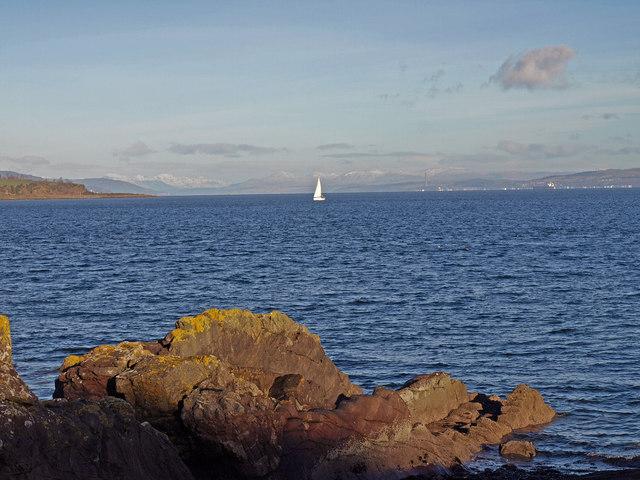}
    \includegraphics[width=.32\linewidth]{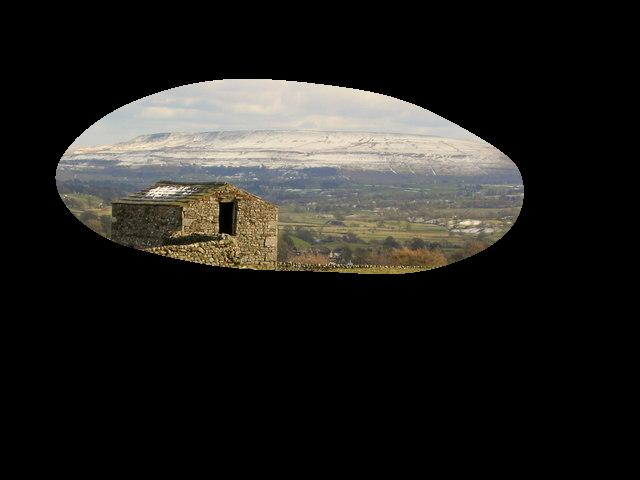}
    \includegraphics[width=.32\linewidth]{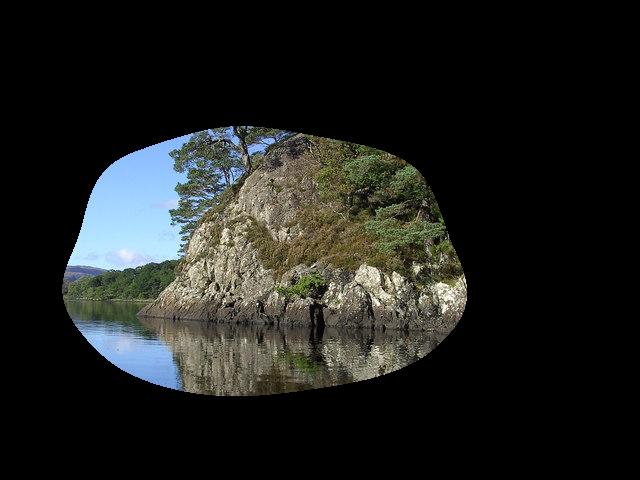}
    \includegraphics[width=.32\linewidth]{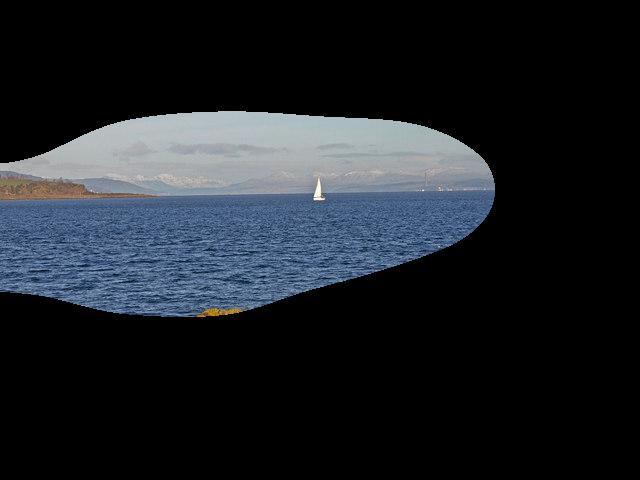}
    \caption{Scenic}
  \end{subfigure}
  \begin{subfigure}{1\linewidth}
    \centering
    \includegraphics[width=.32\linewidth]{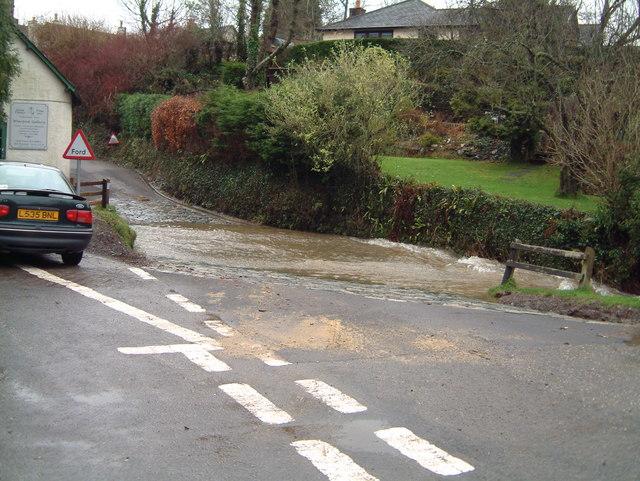}
    \includegraphics[width=.32\linewidth]{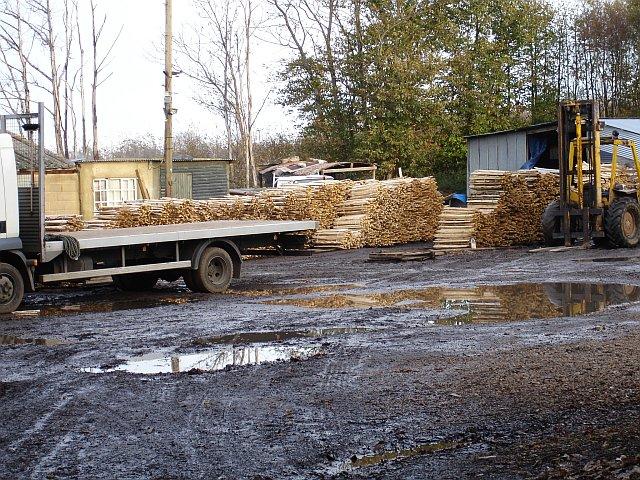}
    \includegraphics[width=.32\linewidth]{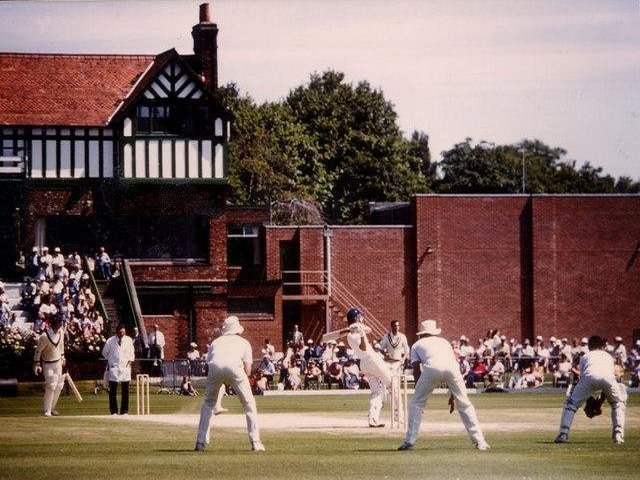}
    \includegraphics[width=.32\linewidth]{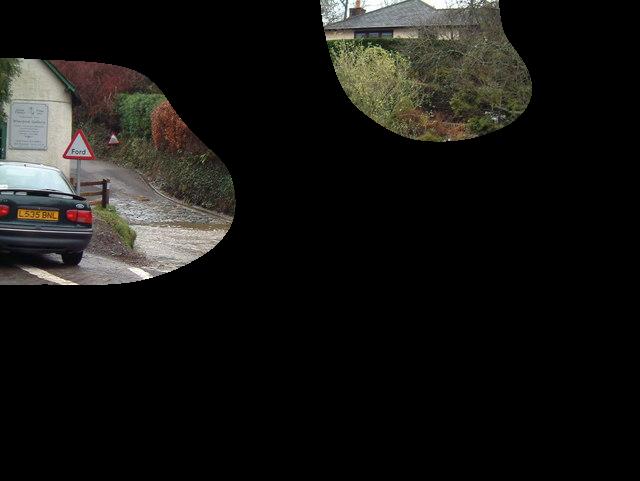}
    \includegraphics[width=.32\linewidth]{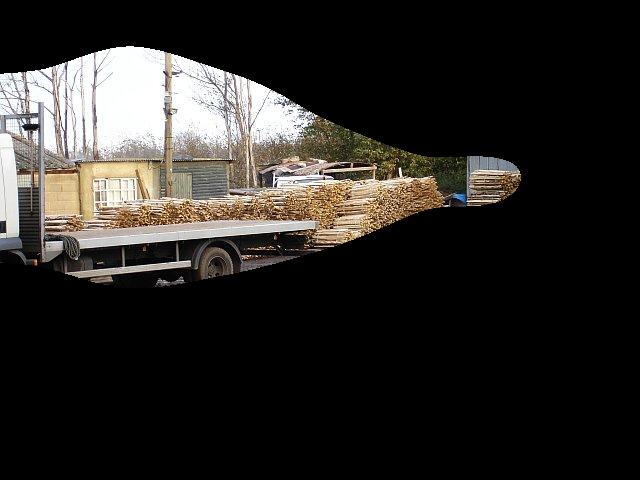}
    \includegraphics[width=.32\linewidth]{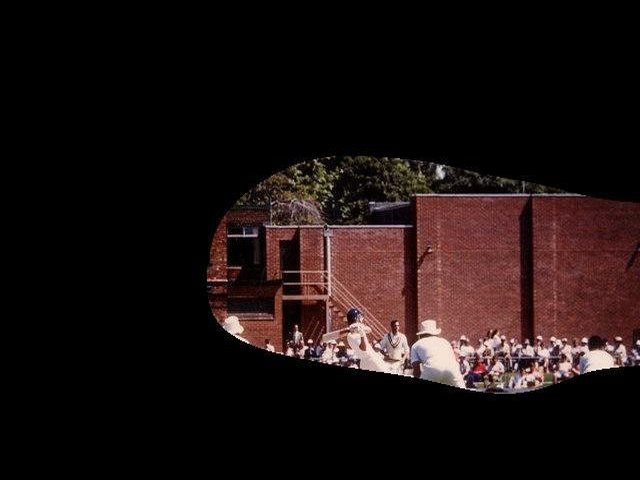}
    \caption{Non-Scenic}
  \end{subfigure}
  
  \caption{Network receptive field analysis. Given an input image (top), the
  output mask (bottom) highlights the region(s) that most significantly impact the maximal
  label assigned by our network.}

  \label{fig:network_vis}
\end{figure}

For additional insight into our scenicness predictions, following Zhou et
al.~\cite{zhou2014object}, we
apply receptive field analysis to highlight the
regions of the image that are most salient in generating the output
distribution. Briefly, the approach computes the differences in
output predictions for a given image with a small (\ie, $7
\times 7$) mask applied. Using a sliding window approach, the prediction 
differences (compared to the unmasked image) are computed on a grid across 
the image. A large difference signifies the masked region plays a significant 
role in the output prediction. This process leads to a saliency map over the 
input image. For visualization purposes, we represent the map
as a binary mask (thresholded at 0.6). 
\figref{network_vis} shows several examples of this
analysis. Each pair of images shows the input and the image regions
with the most contribution to the (high or low) scenicness score. In 
most cases, the receptive fields match the intuition and semantic
analysis of scenicness. Regions containing water, trees, and horizons 
contribute to scenicness, while man-made objects, such as 
buildings and cars, contribute to non-scenicness.

\subsection{Scenicness-Aware Image Cropping}

\begin{figure}%
\includegraphics[width=\columnwidth]{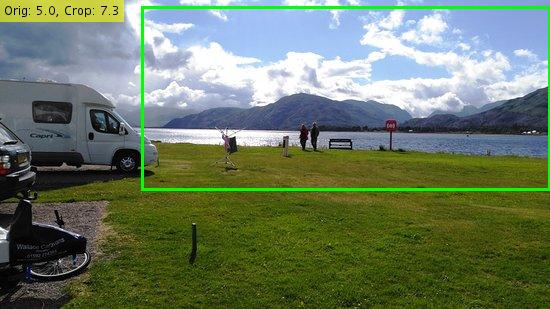} \\
\includegraphics[width=\columnwidth]{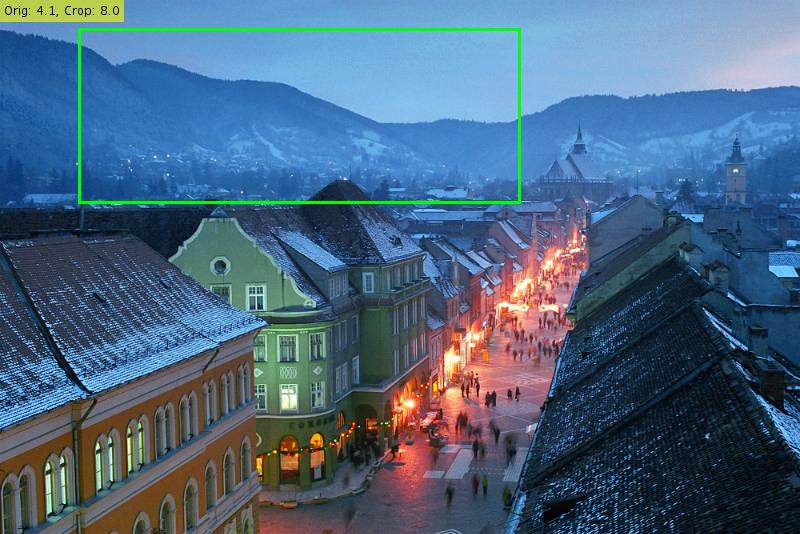} \\
\includegraphics[width=\columnwidth]{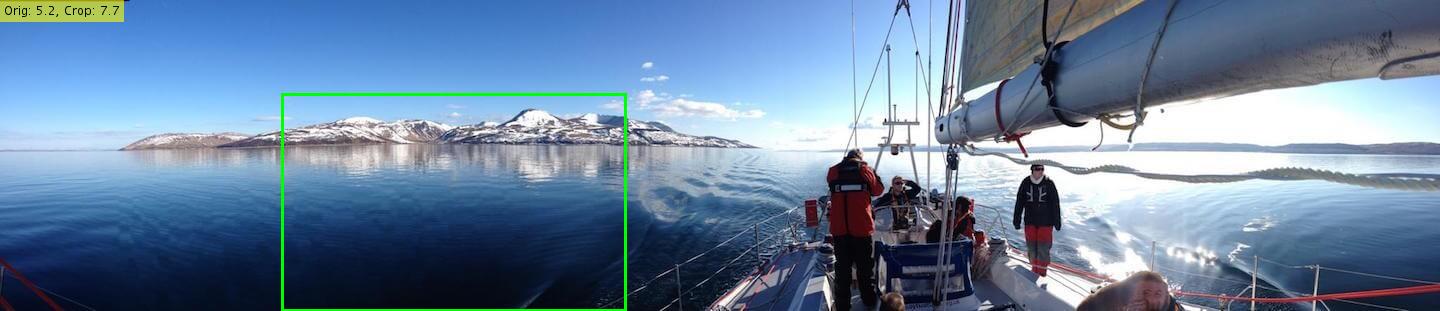}%
\caption{For each image, the green bounding box shows the image crop that maximizes scenicness. 
The predicted scenicness scores for both the entire image and the cropped region are shown in the inset.} 
\label{fig:optcrop}%
\end{figure}

The previous experiment shows that components within a given image contribute differently to
the overall scenicness. For this experiment, we solve for the image crop that maximizes
scenicness. This approach follows the style of previous methods for content-aware 
image processing (\eg, seam carving for image resizing~\cite{avidan2007seam}). We
used constrained Bayesian optimization~\cite{icml2014c2_gardner14} to solve for the position and size
of the maximally scenic image crop, where scenicness is measured as the weighted average prediction
from the {\sc Multinomial} network. Figure~\ref{fig:optcrop} shows representative examples. In some cases,
cropping improved the scenicness scores greatly. For example, in the top image in Figure~\ref{fig:optcrop},
cropping out the vehicles increased the predicted scenicness from 5.0 to 7.3.

\section{Mapping Image Scenicness}
\label{sec:mapping}

\begin{figure*}

  \centering

  \setlength{\myheight}{.16\textwidth}

  \includegraphics[width=\myheight]{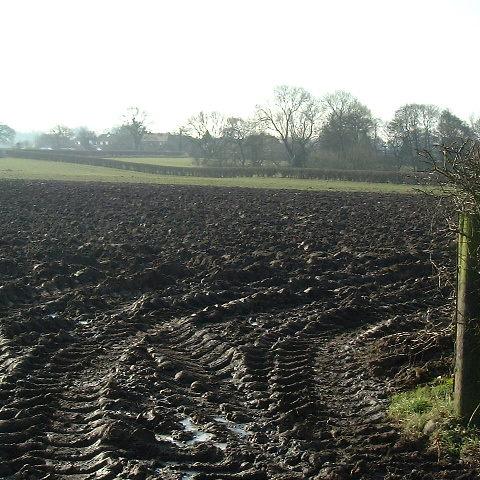}
  \hfill
  \includegraphics[width=\myheight]{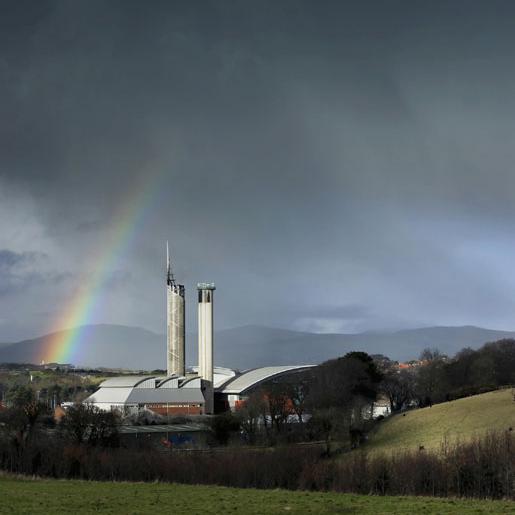}
  \hfill
  \includegraphics[width=\myheight]{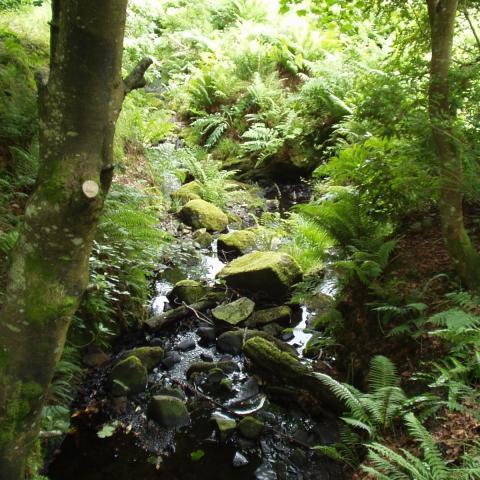}
  \hfill
  \includegraphics[width=\myheight]{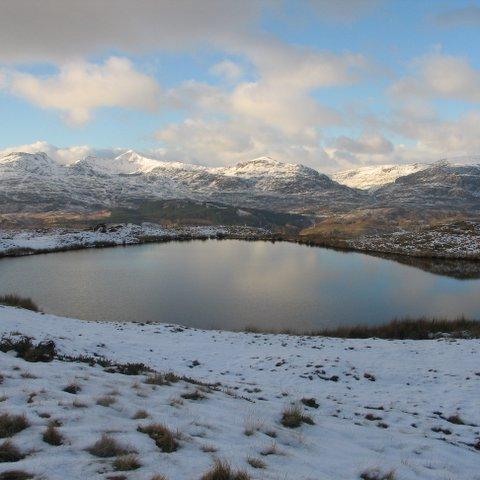}
  \hfill
  \includegraphics[width=\myheight]{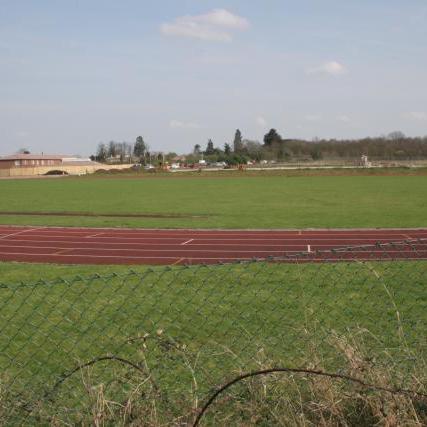}
  \hfill
  \includegraphics[width=\myheight]{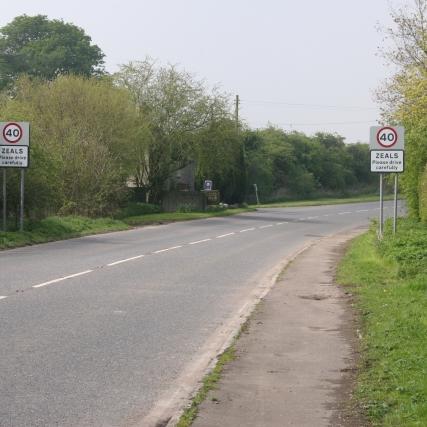}
  
  \includegraphics[width=\myheight]{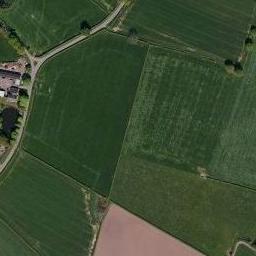}
  \hfill
  \includegraphics[width=\myheight]{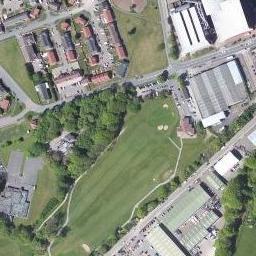}
  \hfill
  \includegraphics[width=\myheight]{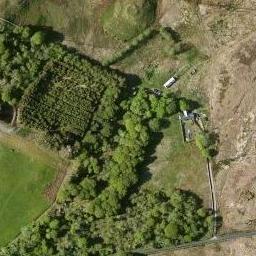}
  \hfill
  \includegraphics[width=\myheight]{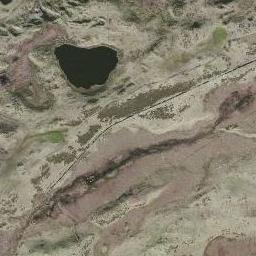}
  \hfill
  \includegraphics[width=\myheight]{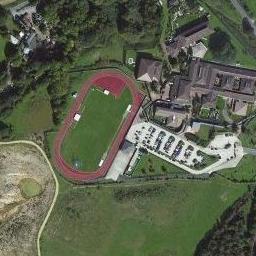}
  \hfill
  \includegraphics[width=\myheight]{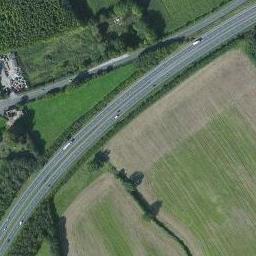}

  \caption{Examples of the co-located ground-level (top) and overhead (bottom)
  image pairs contained in the Cross-View ScenicOrNot (CVSoN) dataset.} 

  \label{fig:cross-view}

\end{figure*}

The previous sections considered scenicness
as a
property of an image. Here, we consider scenicness as a property
of geographic locations and propose a novel approach for estimating
scenicness over a large spatial region. We
extend our approach for single-image estimation to incorporate
overhead imagery.  The result is a dense, high-resolution map that
reflects the scenicness for every location in a region of interest.
Such a map could, for example, be used to provide driving directions
optimized for ``sight seeing''~\cite{quercia2014shortest,runge2016no} or suggest
places to go for a walk~\cite{quercia2015digital}. 

We consider geotagged images as noisy samples of the underlying
geospatial scenicness function.  The challenge is that ground-level
imagery is sparsely distributed,
especially away from major urban areas and tourist
attractions.  This means that methods which estimate maps using only
ground-level
imagery~\cite{arietta2014city,lorenzo2015safety,xie2011im2map}
typically generate either low-resolution or noisy maps.

To deal with the problem of interpolating sparse examples over large
spatial regions, we apply a cross-view training and mapping
approach. Cross-view
methods~\cite{lin2015learning,workman2015wide,zhai2017predicting}
incorporate both ground-level and overhead viewpoints and take advantage
of the fact that, while ground-level images are spatially sparse,
overhead imagery is available at a high-resolution in most locations.
Jointly reasoning about ground-level and overhead imagery has become
popular in recent years. Luo et al.~\cite{luo2008event} use overhead 
imagery to perform better event recognition by fusing
complementary views. Lin et al.~\cite{lin2013cross,lin2015learning}
introduce the problem of cross-view geolocalization, where an overhead 
image reference database is used to support ground-level image
localization by learning a feature mapping between the two viewpoints.
Workman et al.~\cite{workman2015geocnn,workman2015wide} study the
geo-dependence of image features and propose a cross-view training
approach.

To support these efforts, we extend the ScenicOrNot (SoN) dataset to
incorporate overhead images. Specifically, for each geotagged,
ground-level SoN image, we obtained a $256 \times 256$ orthorectified
overhead image centered at that location from Bing Maps (zoom level 16,
which is $\sim$2.4 meters/pixel).  \figref{cross-view} shows
co-located pairs of ground-level and overhead images from the Cross-View
ScenicOrNot (CVSoN) dataset. The dataset is available at
our project website.\textsuperscript{\ref{projectsite}}

\subsection{Cross-View Mapping}

To predict the scenicness of an overhead image even though labeled
overhead images are not available, we apply a cross-view training
strategy; instead of predicting the scenicness of the overhead image, we
predict the scenicness of a ground-level image captured at the same
location.  
We use the same network architecture and training methods
as with the ground-level network, with two changes: (1) overhead 
(instead of ground-level) images are used as input and (2) the weights
are initialized with those learned from the ground-level network.
Similar to our ground-level network, after training, the output of
this overhead image network is a distribution over scenicness ratings.

While using overhead images as input may address the issue of sparse
spatial coverage of ground-level imagery, an overhead-only network
may miss, for example, scenic views hidden amongst dense urban areas.
To address this issue, we introduce a novel variant to the cross-view
approach for combining ground-level and overhead imagery to
estimate the scenicness of a query location. This is similar to our
framework for estimating geospatial
functions~\cite{workman2017unified}.

\begin{figure}
  \centering
  \includegraphics[clip,trim=.45cm 1cm .6cm .3cm,width=\linewidth]{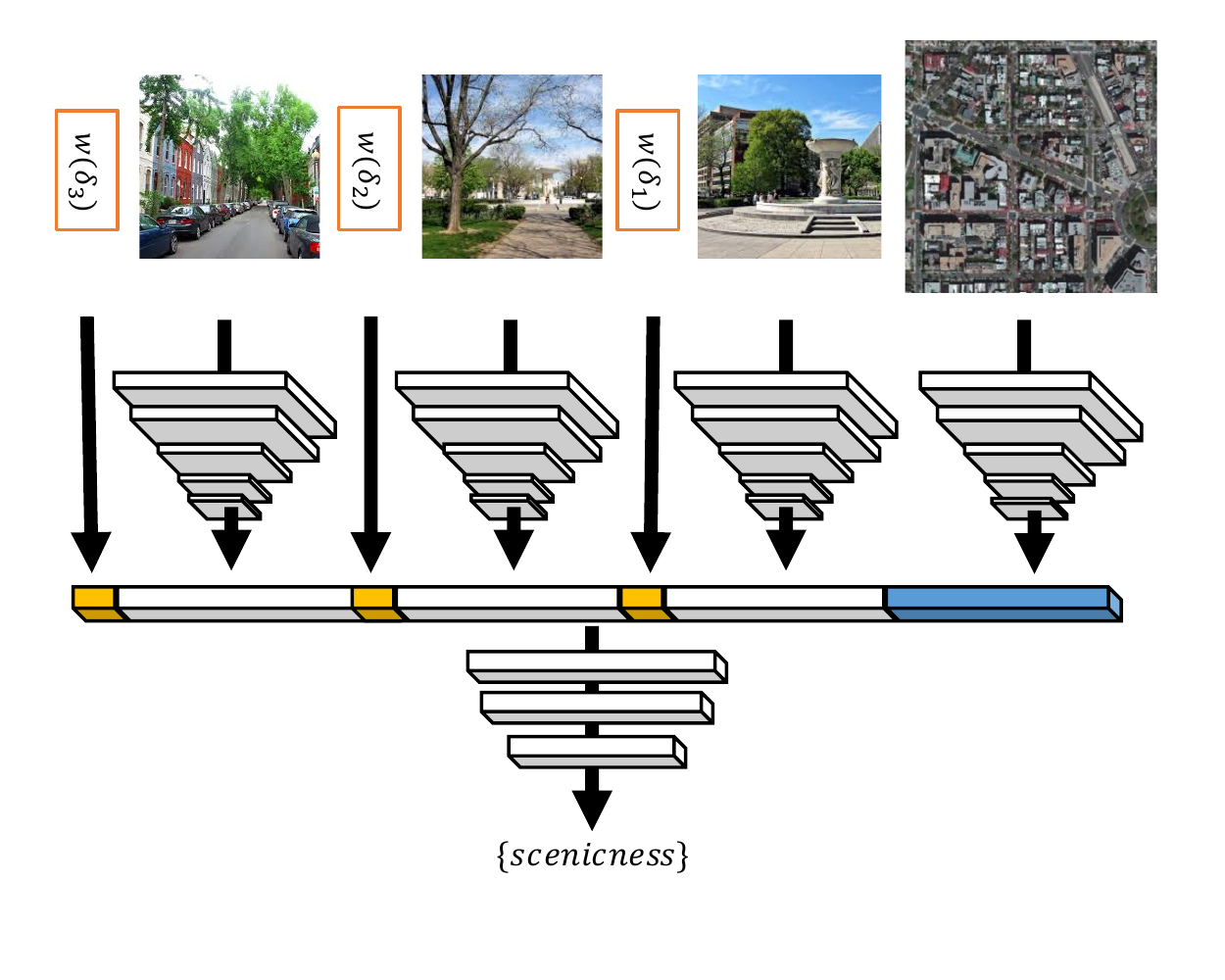}
  \caption{The architecture for our hybrid approach to cross-view mapping.}
  \label{fig:cvh}
\end{figure}

\figref{cvh} shows an overview of our hybrid cross-view
approach.  For a given query location, $q$, consider the co-located
overhead image, set of the $k$ closest ground-level images, and the
distances of the ground-level images to the query location,
$\{\delta_1,\delta_2,\dots, \delta_k\}$.  For the images, we can
compute scenicness features using the existing ground and overhead 
networks.  For the hybrid approach, we learn and predict scenicness
from the fused features (overhead image features, ground-level features,
weighted distances) using a small feed-forward network, with three 
hidden layers containing 100, 50, and 25 neurons, respectively.  The
activation function on the internal nodes is the hyperbolic tangent
sigmoid. The network weights are regularized using an $L_2$ loss with a
weight of $0.5$. The
output is the predicted distribution of ratings for a ground-level
image taken at the input location.  We refer to this as the
\emph{Cross-View Hybrid (CVH)} network.

\subsection{Mapping the Scenicness of Great Britain}

To evaluate CVH, the CVSoN dataset is divided as before, with the same
1,413 ground-level images (with at least 10 ratings) held out for
testing.  For CVH, the test input includes the co-located overhead image. We
compare against two baseline methods for constructing dense maps of
visual properties: 
\begin{itemize}[itemsep=0pt]
  \item {\em 1NN}: return the prediction from the ground-level image closest to
    the query location; and

  \item {\em LWA}: return the locally weighted average prediction of neighboring 
	ground-level images with a Gaussian kernel
    ($\sigma=0.01$ degrees).
  
\end{itemize}
To compare our methods, we formulate a binary classification task to
determine if a given test image is above or below a scenicness rating
of 7. \tblref{mapping} shows the results for each method as the area
under the curve (AUC) of the ROC curve computed from the output
distributions.  The results show that including orthographic overhead 
imagery improves the resulting predictions.  

\begin{table}
  \centering
  \caption{Comparison of mapping strategies.}
  \begin{tabular}{@{}cccc@{}}
    \toprule
    \textbf{Method} & 1NN & LWA & CVH \\
    \hline
    \textbf{AUC} & 64.38\% & 66.86\% & 68.51\% \\
    \bottomrule
  \end{tabular}
  \label{tbl:mapping}
\end{table}

\begin{figure}
  \centering
 
  \setlength{\tabcolsep}{1pt}
	\setlength{\myheight}{.24\linewidth}
  \begin{tabular}{cccc}
    ROI & 1NN & LWA & CVH \\ 
    \includegraphics[clip,trim=0cm 1cm 1cm 0cm,width=\myheight]{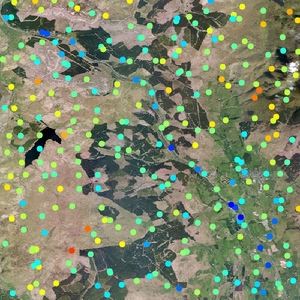} &
    \includegraphics[clip,trim=0cm 1cm 1cm 0cm,width=\myheight]{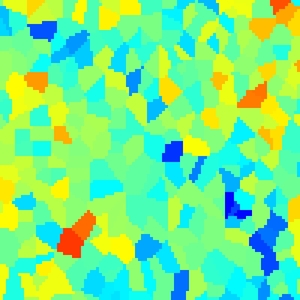} &
    \includegraphics[clip,trim=0cm 1cm 1cm 0cm,width=\myheight]{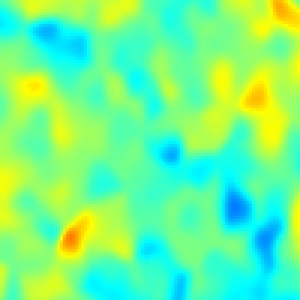} &
    \includegraphics[clip,trim=0cm 1cm 1cm 0cm,width=\myheight]{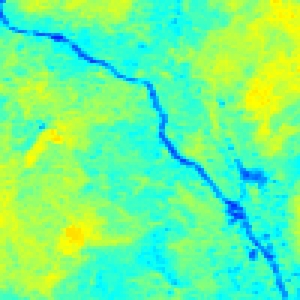} \\
    
    \includegraphics[clip,trim=0cm 1cm 1cm 0cm,width=\myheight]{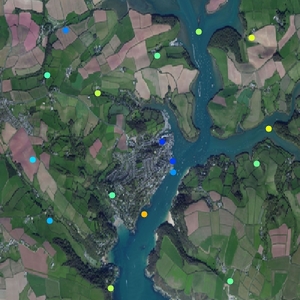} &
    \includegraphics[clip,trim=0cm 1cm 1cm 0cm,width=\myheight]{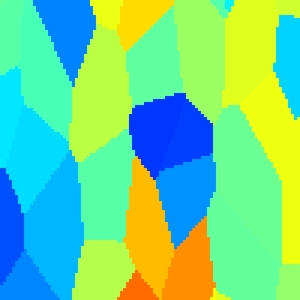} &
    \includegraphics[clip,trim=0cm 1cm 1cm 0cm,width=\myheight]{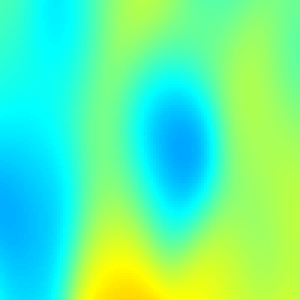} &
    \includegraphics[clip,trim=0cm 1cm 1cm 0cm,width=\myheight]{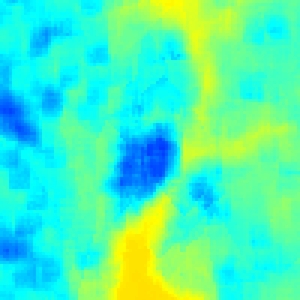} \\
    
    \includegraphics[clip,trim=0cm 1cm 1cm 0cm,width=\myheight]{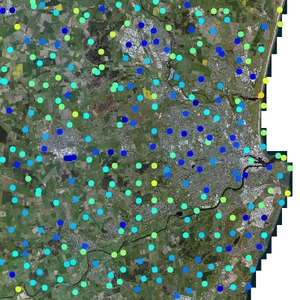} &
    \includegraphics[clip,trim=0cm 1cm 1cm 0cm,width=\myheight]{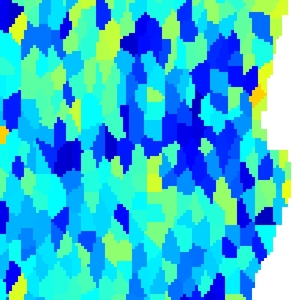} &
    \includegraphics[clip,trim=0cm 1cm 1cm 0cm,width=\myheight]{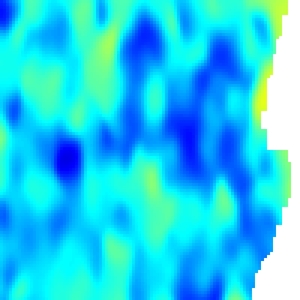} &
    \includegraphics[clip,trim=0cm 1cm 1cm 0cm,width=\myheight]{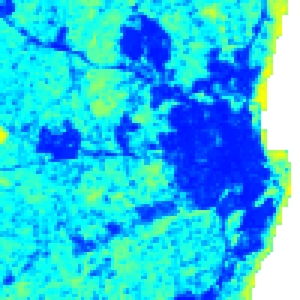} \\
		
		   \includegraphics[clip,trim=0cm 1cm 1cm 0cm,width=\myheight]{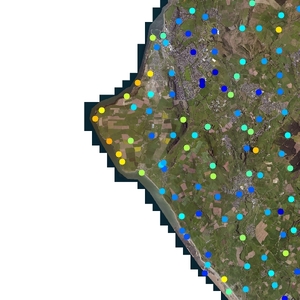} &
    \includegraphics[clip,trim=0cm 1cm 1cm 0cm,width=\myheight]{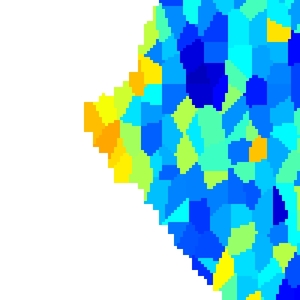} &
    \includegraphics[clip,trim=0cm 1cm 1cm 0cm,width=\myheight]{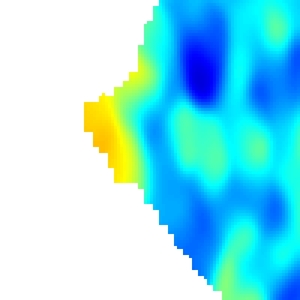} &
    \includegraphics[clip,trim=0cm 1cm 1cm 0cm,width=\myheight]{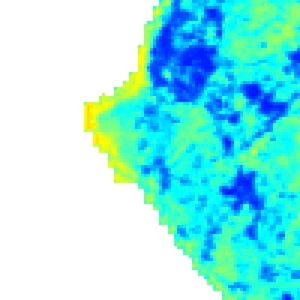} \\
		
		   \includegraphics[clip,trim=0cm 1cm 1cm 0cm,width=\myheight]{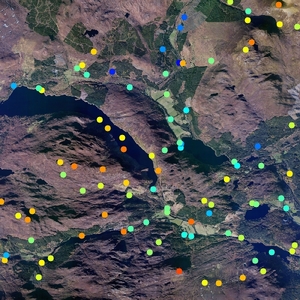} &
    \includegraphics[clip,trim=0cm 1cm 1cm 0cm,width=\myheight]{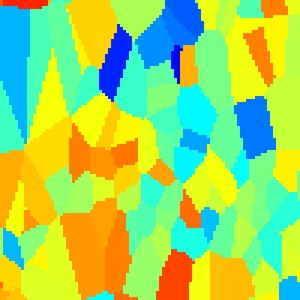} &
    \includegraphics[clip,trim=0cm 1cm 1cm 0cm,width=\myheight]{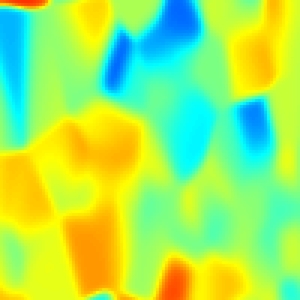} &
    \includegraphics[clip,trim=0cm 1cm 1cm 0cm,width=\myheight]{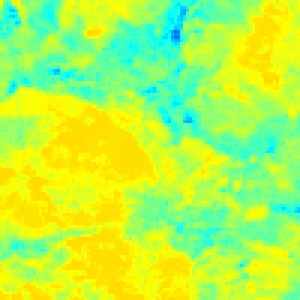} \\

  \end{tabular}

  \caption{Scenicness maps. The first
    column shows an overhead image where dots correspond to geotagged ground-level
    imagery, colored by average scenicness rating (warmer colors correspond to more scenic images). The remaining columns show false-color images that
    reflect the average scenicness predicted by each method.}

  \label{fig:map}
\end{figure}

These results are supported qualitatively in \figref{map}, which
shows scenicness maps for several regions around Great Britain.
We observe that by including overhead imagery we are able to construct a
significantly more accurate map than purely interpolating scenicness
estimates obtained from ground-level images alone. The maps
created using only ground-level images (\eg, 1NN, LWA) are
susceptible to both underprediction (\eg, no nearby scenic ground-level
images) and overprediction (\eg, a single nearby scenic image with
a narrow field of view). On the other hand, the cross-view approach can
be more robust against these types of mispredictions due to 
effectively averaging across many images (by marginalizing through the overhead imagery),
not just those in the nearby area.

\section{Conclusions}

We explored the concept of natural beauty as it pertains to outdoor
imagery. Using a dataset containing hundreds of thousands of
ground-level images rated by humans, we showed it is possible to
quantify scenicness, from both ground-level and overhead viewpoints.
To our knowledge, this is the first time a combination of overhead and
geotagged ground-level imagery has been used to map the scenicness of
a region. The resulting maps are higher-resolution than those
constructed by previous approaches and can be quickly computed.  Such
methods have significant practical importance to many areas,
including: tourism, transportation routing, and environmental monitoring.  

\ificcvfinal
\paragraph*{Acknowledgments}
We thank our colleagues, Tawfiq Salem and Zachary Bessinger, for their
help, and Robert Pless and Mirek Truszczynski for their insightful
advice. We gratefully acknowledge the support of NSF CAREER grant
IIS-1553116 and a Google Faculty Research Award.
\fi

{\small
\bibliographystyle{ieee}
\bibliography{biblio}
}

\input{supplemental}

\end{document}

%% file: supplemental.tex
\newpage
\null
\vskip .375in
\twocolumn[{%
  \begin{center}
    \textbf{\Large Supplemental Material : Understanding and Mapping Natural Beauty}
  \end{center}
  \vspace*{24pt}
}]
\setcounter{page}{1}
\setcounter{section}{0}
\setcounter{equation}{0}
\setcounter{figure}{0}
\setcounter{table}{0}
\makeatletter
\renewcommand{\theequation}{S\arabic{equation}}
\renewcommand{\thefigure}{S\arabic{figure}}

This document contains additional details and experiments related to
our methods. 

\section{Exploring Image Scenicness}

\subsection{Uncertainty in Scenicness Scores}

\begin{figure*}
  \centering
  \includegraphics[clip,trim=1.1cm .4cm 2.05cm .6cm,width=1\linewidth]{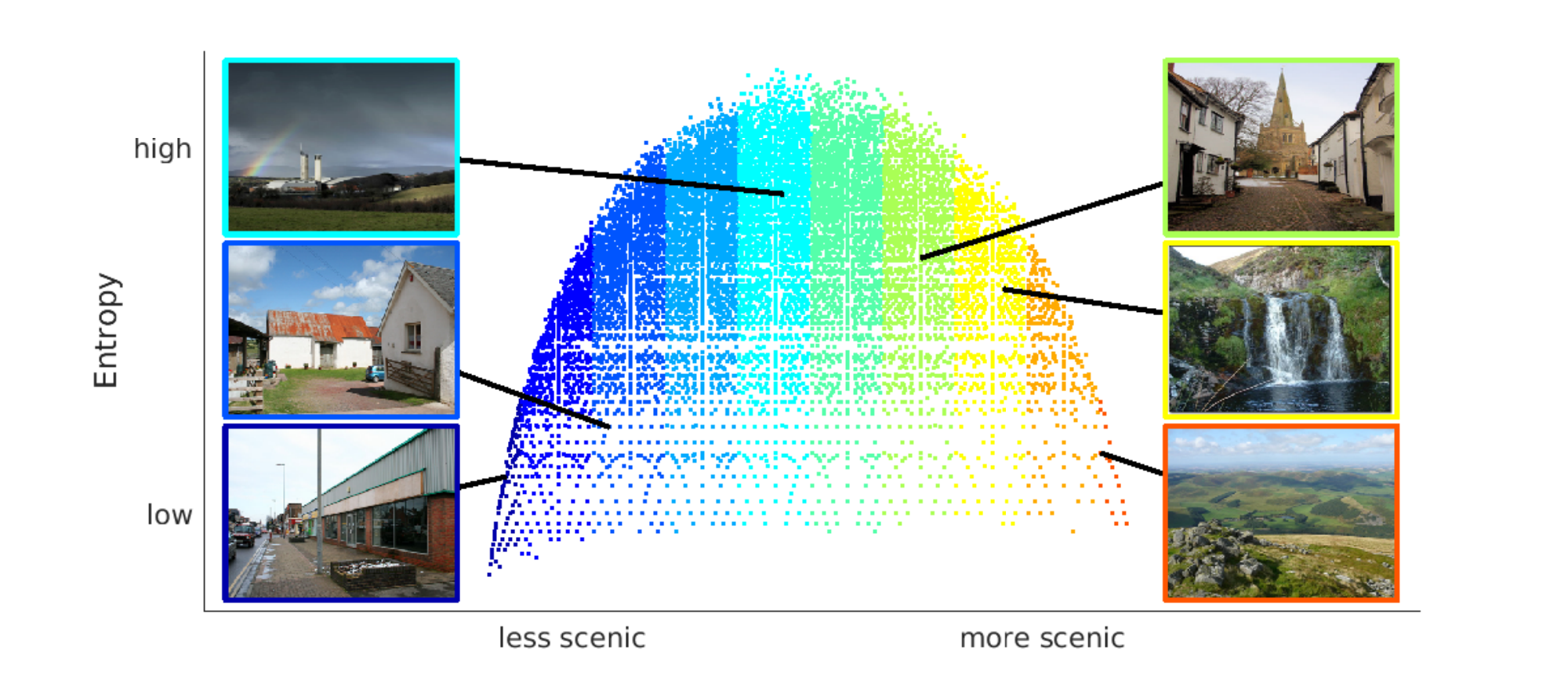}
  \caption{A scatter plot of per-image entropy vs. scenicness,
  computed from human ratings, for each image in the ScenicOrNot
  dataset.} 
  \label{fig:s_entropy}
\end{figure*}

\figref{s_entropy} shows a scatter plot of Shannon entropy, computed per
image from ratings, versus scenicness, for all images in the ScenicOrNot
(SoN) dataset.  Each dot represents an image and is colored by the
average beauty score. Unsurprisingly, the images that are rated the
least and most scenic have higher consistency in their scenicness
ratings. 

\subsection{Image Captions}
\begin{figure}
  \centering
  \includegraphics[clip,trim=1.5cm .5cm .8cm 1.6cm,width=\linewidth]{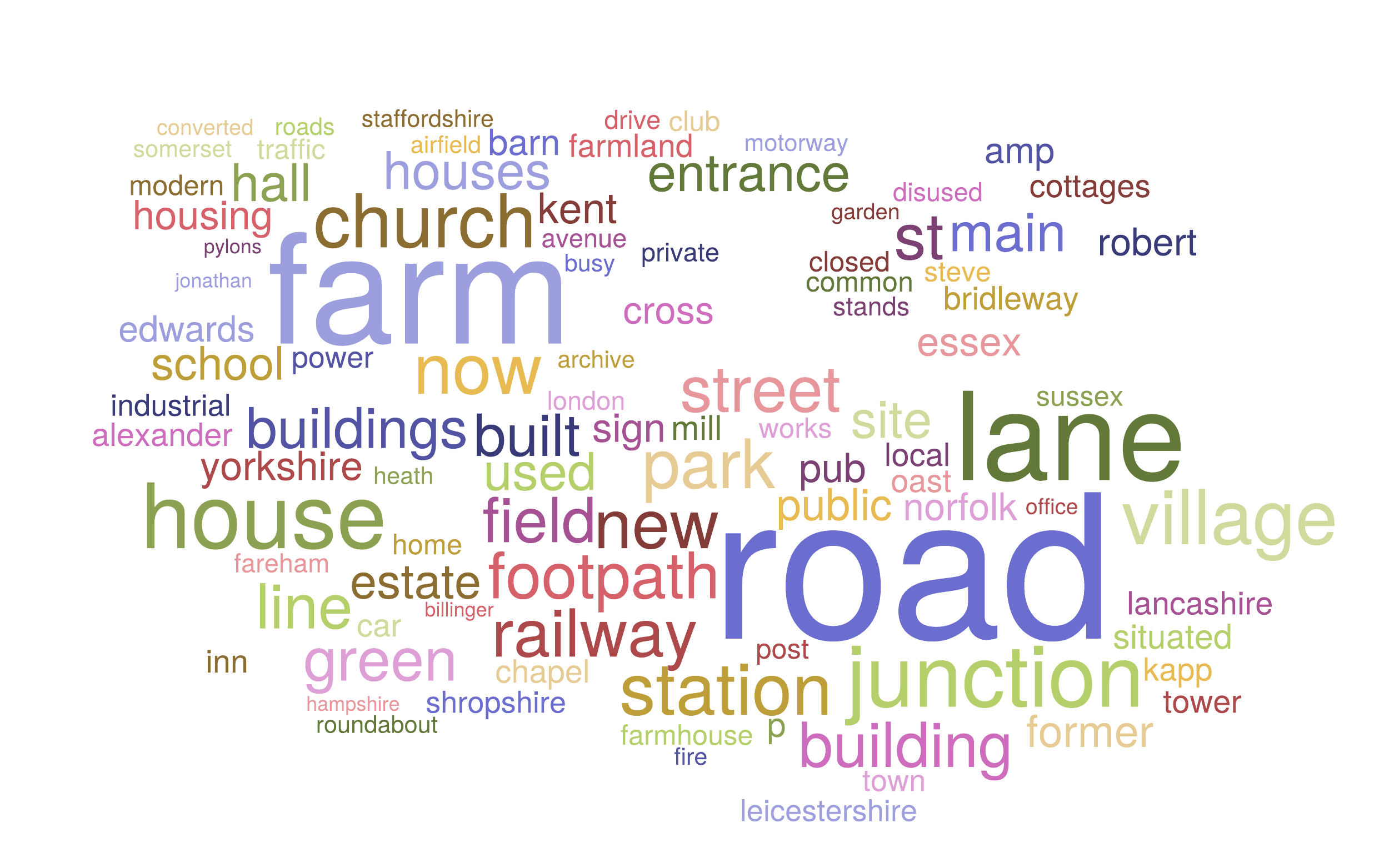}
  \caption{The word cloud depicts the relative frequency of title and
  caption terms found in non-scenic images from the ScenicOrNot
  dataset.}
  \label{fig:s_wordcloud}
\end{figure}

\figref{s_wordcloud} shows a word cloud of the most frequent 100
extracted terms from captions and titles associated with non-scenic
images in SoN (average rating below 3.0), where the size of the word
represents the relative frequency.  Example terms that are negatively
correlated with scenicness include ``road'', ``lane'', ``house'', and
``railway''.

\section{Experiments}

\subsection{Scenicness-Aware Image Cropping}

\begin{figure*}
  \centering
 
  \setlength{\tabcolsep}{1pt}
  \begin{tabular}{cc}
    \includegraphics[height=.3\linewidth]{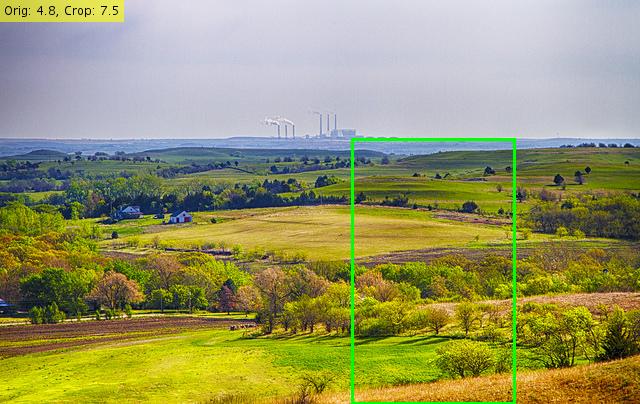} &
    \includegraphics[height=.3\linewidth]{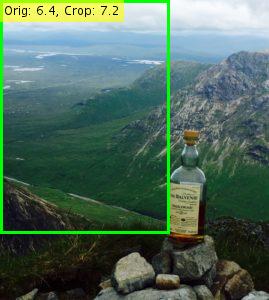} \\ 
    
    \includegraphics[height=.3\linewidth]{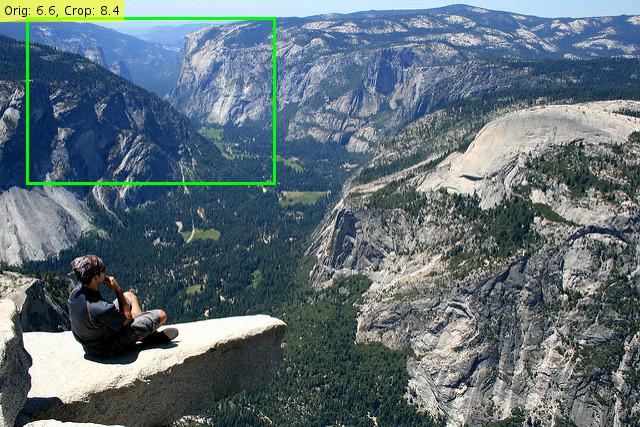} &
    \includegraphics[height=.3\linewidth]{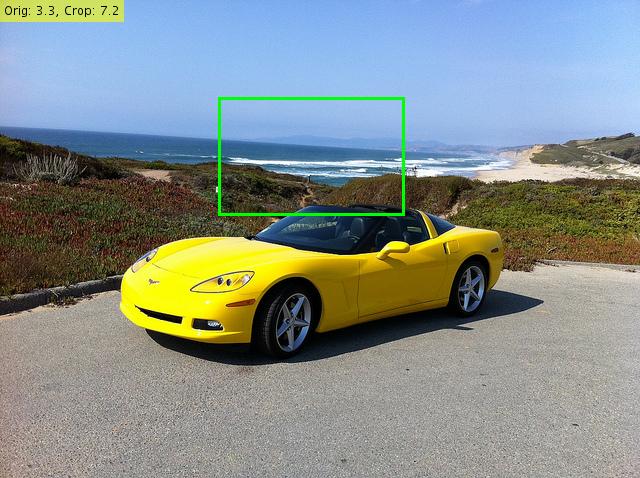} \\
    
    \includegraphics[height=.3\linewidth]{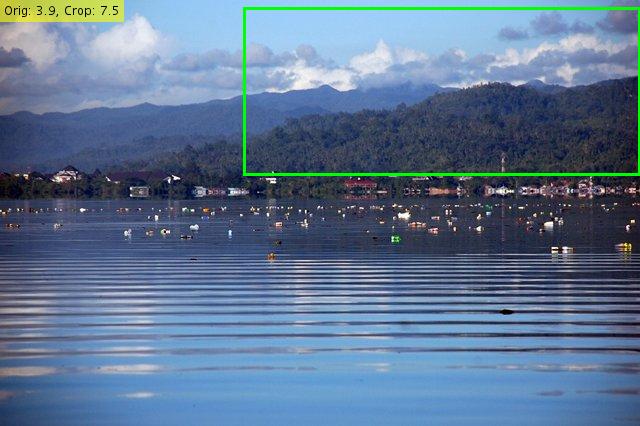} & 
    \includegraphics[height=.3\linewidth]{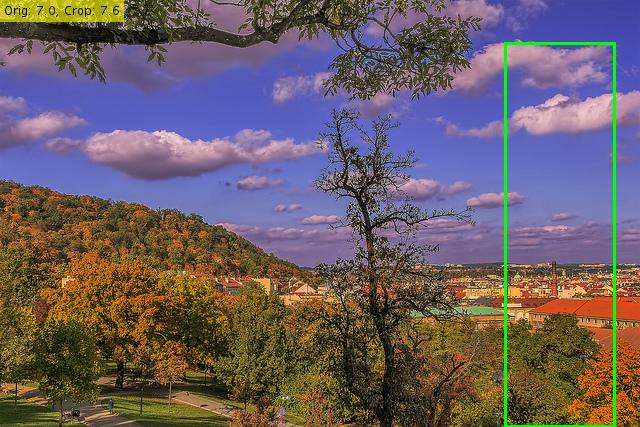} \\
    
    \includegraphics[height=.3\linewidth]{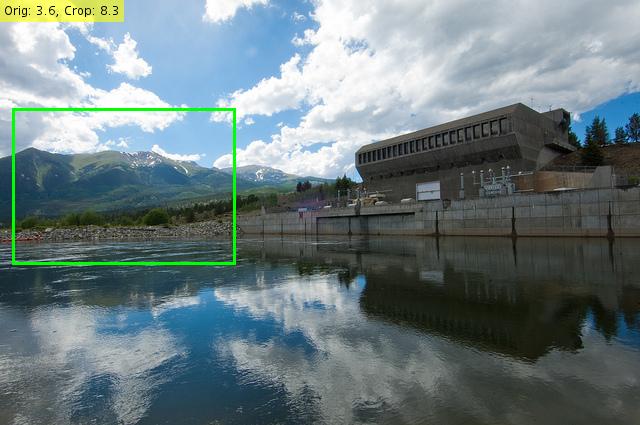} &
    \includegraphics[height=.3\linewidth]{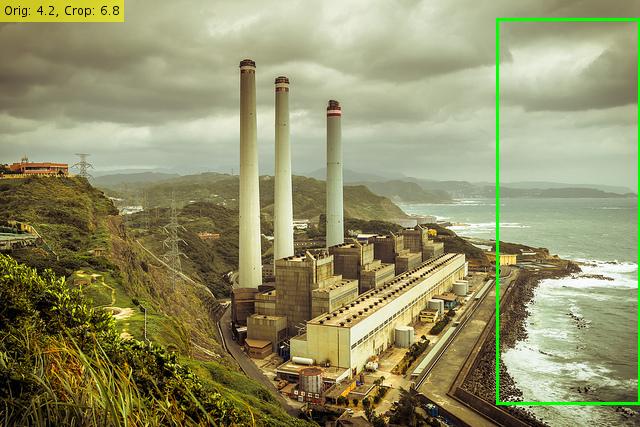} \\
  \end{tabular}
  
  \caption{Scenicness-aware image processing. For each image, the
  green bounding box shows the image crop that maximizes scenicness.
  The predicted scenicness scores for both the entire image and the
  cropped region are shown in the inset.} 

  \label{fig:s_optcrop}
\end{figure*}

\figref{s_optcrop} shows additional examples of applying our methods for
scenicness-aware image processing. In each image, the inset shows the
change in scenicness from the full image to the cropped image. 

\subsection{Mapping Scenicness}

\begin{figure*}
  \centering
 
  \begin{tabular}{cc}
    ROI & CVH \\
    \includegraphics[width=.33\linewidth]{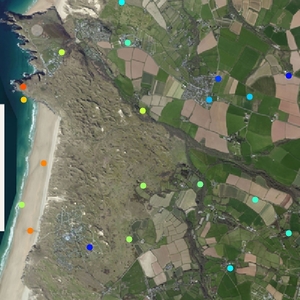} &
    \includegraphics[width=.33\linewidth]{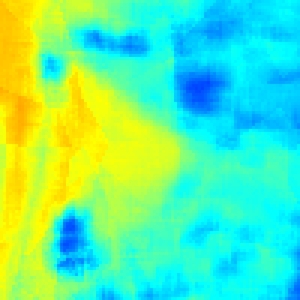} \\
    
    \includegraphics[width=.33\linewidth]{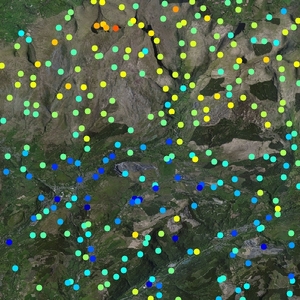} &
    \includegraphics[width=.33\linewidth]{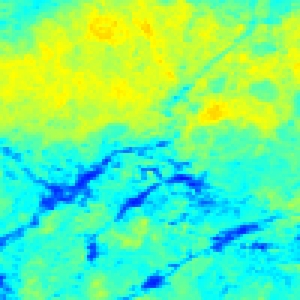} \\
    
    \includegraphics[width=.33\linewidth]{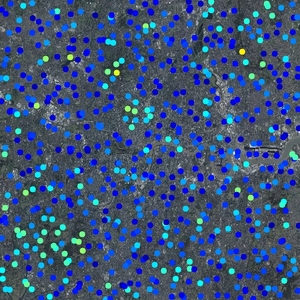} &
    \includegraphics[width=.33\linewidth]{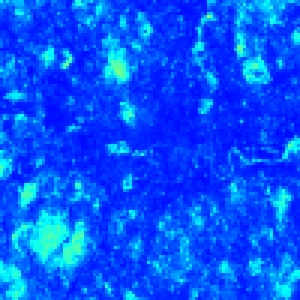} \\
    
  \end{tabular}

  \caption{Varying spatial resolutions. The first column shows an
  overhead image where dots correspond to geotagged ground-level
  imagery, colored by average scenicness rating (warmer colors
  correspond to more scenic images). The second column shows a
  false-color image that reflects the average scenicness predicted by
  our method.}

  \label{fig:s_fine}
\end{figure*}

\begin{figure*}
  \centering
 
  \setlength{\tabcolsep}{1pt}
  \begin{tabular}{cccc}
    ROI & 1NN & LWA & CVH \\
    \includegraphics[width=.15\linewidth]{mapping/hills/roi} &
    \includegraphics[width=.15\linewidth]{mapping/hills/1nn} &
    \includegraphics[width=.15\linewidth]{mapping/hills/lwa} &
    \includegraphics[width=.15\linewidth]{mapping/hills/all_net} \\
    
    \includegraphics[width=.15\linewidth]{mapping/beach/roi} &
    \includegraphics[width=.15\linewidth]{mapping/beach/1nn} &
    \includegraphics[width=.15\linewidth]{mapping/beach/lwa} &
    \includegraphics[width=.15\linewidth]{mapping/beach/all_net} \\
    
    \includegraphics[width=.15\linewidth]{mapping/city/roi} &
    \includegraphics[width=.15\linewidth]{mapping/city/1nn} &
    \includegraphics[width=.15\linewidth]{mapping/city/lwa} &
    \includegraphics[width=.15\linewidth]{mapping/city/all_net} \\
    
    \includegraphics[width=.15\linewidth]{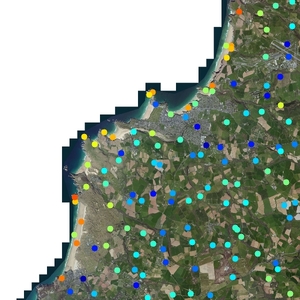} &
    \includegraphics[width=.15\linewidth]{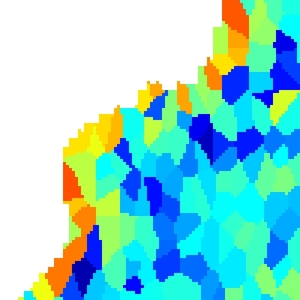} &
    \includegraphics[width=.15\linewidth]{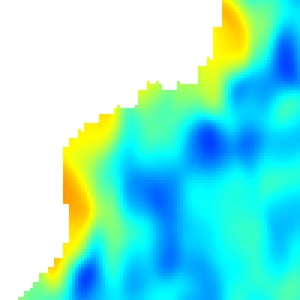} &
    \includegraphics[width=.15\linewidth]{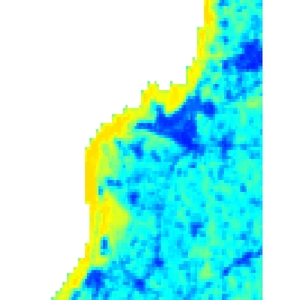} \\
    
    \includegraphics[width=.15\linewidth]{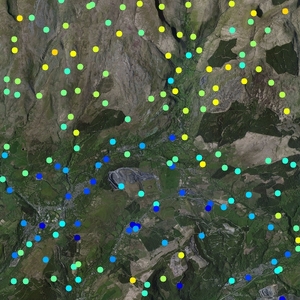} &
    \includegraphics[width=.15\linewidth]{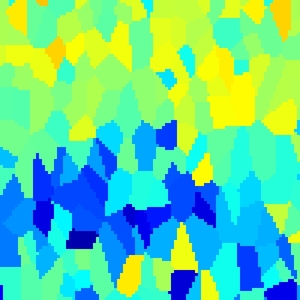} &
    \includegraphics[width=.15\linewidth]{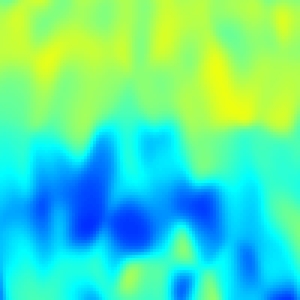} &
    \includegraphics[width=.15\linewidth]{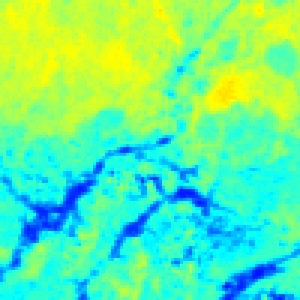} \\
    
    \includegraphics[width=.15\linewidth]{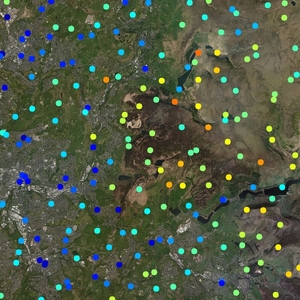} &
    \includegraphics[width=.15\linewidth]{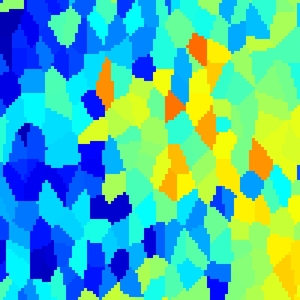} &
    \includegraphics[width=.15\linewidth]{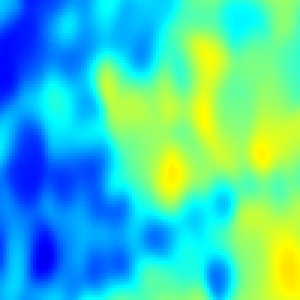} &
    \includegraphics[width=.15\linewidth]{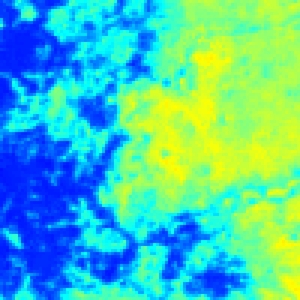} \\
    
    \includegraphics[width=.15\linewidth]{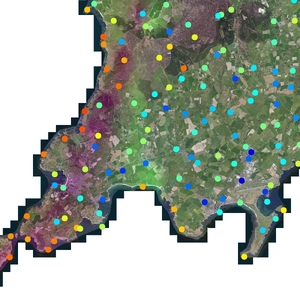} &
    \includegraphics[width=.15\linewidth]{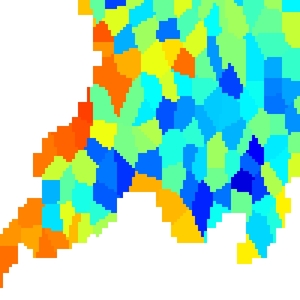} &
    \includegraphics[width=.15\linewidth]{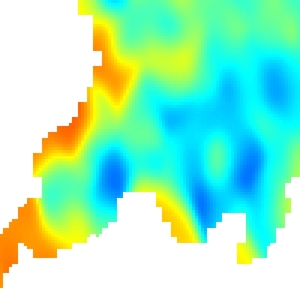} &
    \includegraphics[width=.15\linewidth]{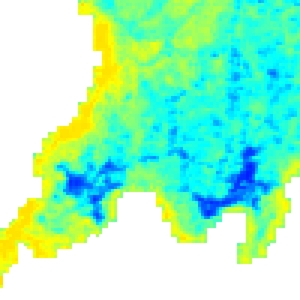} \\

  \end{tabular}

  \caption{Scenicness maps. The first column shows an overhead image
  where dots correspond to geotagged ground-level imagery, colored by
  average scenicness rating (warmer colors correspond to more scenic
  images). The remaining columns show false-color images that reflect
  the average scenicness predicted by each method.}

  \label{fig:s_map3}
\end{figure*}

\begin{figure*}
  \centering
 
  \setlength{\tabcolsep}{1pt}
  \begin{tabular}{cccc}
    ROI & 1NN & LWA & CVH \\
    \includegraphics[width=.15\linewidth]{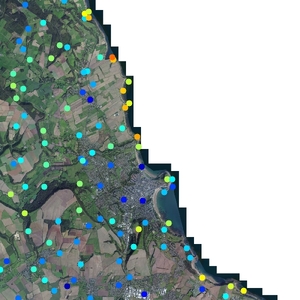} &
    \includegraphics[width=.15\linewidth]{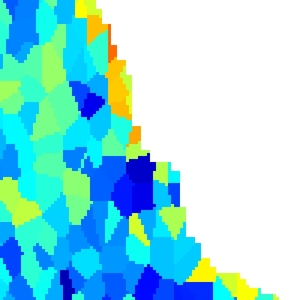} &
    \includegraphics[width=.15\linewidth]{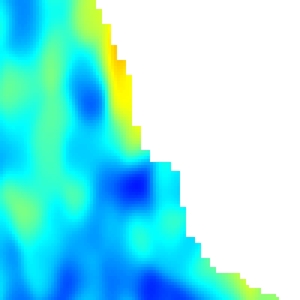} &
    \includegraphics[width=.15\linewidth]{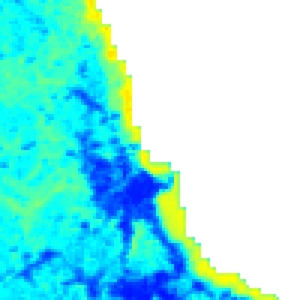} \\
    
    \includegraphics[width=.15\linewidth]{mapping/region_6/roi} &
    \includegraphics[width=.15\linewidth]{mapping/region_6/1nn} &
    \includegraphics[width=.15\linewidth]{mapping/region_6/lwa} &
    \includegraphics[width=.15\linewidth]{mapping/region_6/all_net} \\
    
    \includegraphics[width=.15\linewidth]{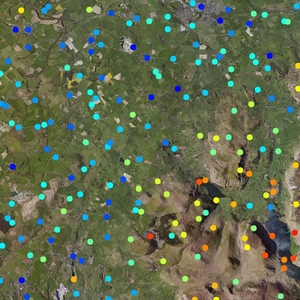} &
    \includegraphics[width=.15\linewidth]{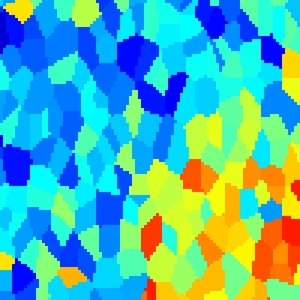} &
    \includegraphics[width=.15\linewidth]{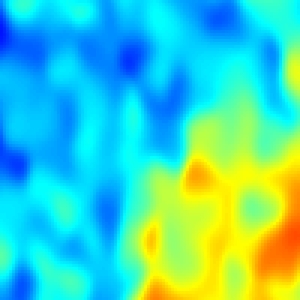} &
    \includegraphics[width=.15\linewidth]{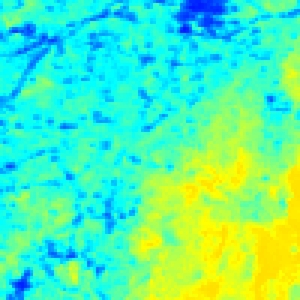} \\
    
    \includegraphics[width=.15\linewidth]{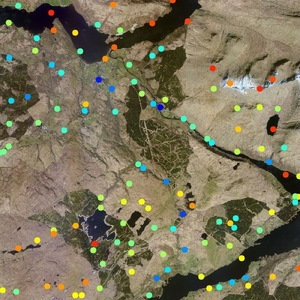} &
    \includegraphics[width=.15\linewidth]{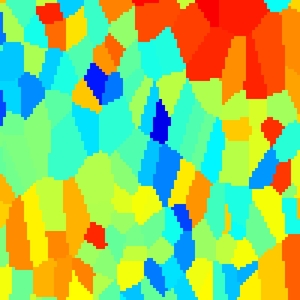} &
    \includegraphics[width=.15\linewidth]{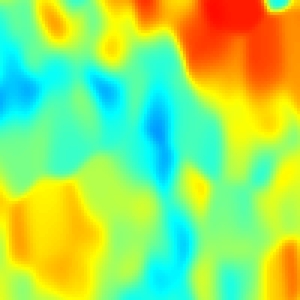} &
    \includegraphics[width=.15\linewidth]{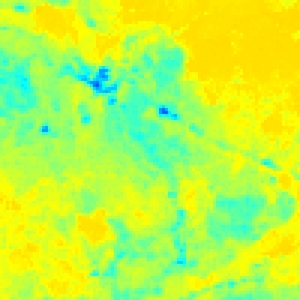} \\
    
    \includegraphics[width=.15\linewidth]{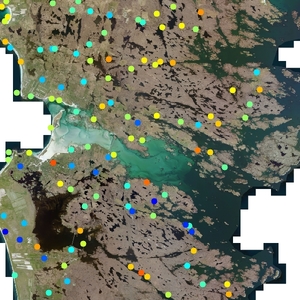} &
    \includegraphics[width=.15\linewidth]{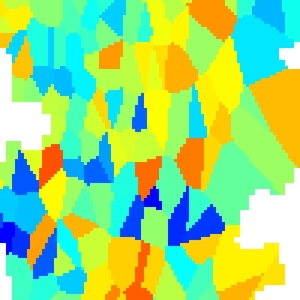} &
    \includegraphics[width=.15\linewidth]{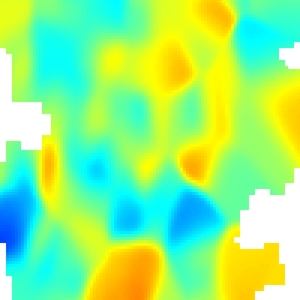} &
    \includegraphics[width=.15\linewidth]{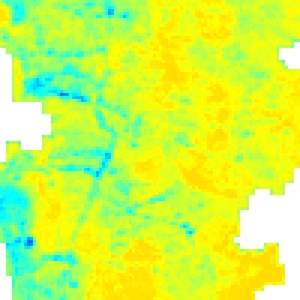} \\
    
    \includegraphics[width=.15\linewidth]{mapping/region_10/roi} &
    \includegraphics[width=.15\linewidth]{mapping/region_10/1nn} &
    \includegraphics[width=.15\linewidth]{mapping/region_10/lwa} &
    \includegraphics[width=.15\linewidth]{mapping/region_10/all_net} \\

  \end{tabular}

  \caption{Scenicness maps. The first column shows an overhead image
  where dots correspond to geotagged ground-level imagery, colored by
  average scenicness rating (warmer colors correspond to more scenic
  images). The remaining columns show false-color images that reflect
  the average scenicness predicted by each method.}

  \label{fig:s_map4}
\end{figure*}

Our method is applicable for generating maps of scenicness at widely
varying spatial scales. \figref{s_fine} shows three examples of
fine-grained high-resolution maps of scenicness from regions of
different sizes around Great Britain. \figref{s_fine} (top) shows a
stretch of coast near Holywell, a coastal village in north Cornwall,
England.  \figref{s_fine} (middle) shows a map centered over Caehopkin,
a village in Powys, Wales. Caehopkin sits between Abercraf and
Coelbren in the Swansea Valley and lies on the border of the Brecon
Beacons National Park to the north. \figref{s_fine} (bottom) shows a map
of Greater London.  \figref{s_map3} and \figref{s_map4} show additional
maps of scenicness computed using our method, alongside several
baseline approaches. In all cases, {\em Cross-View Hybrid (CVH)}
combines overhead imagery and nearby ground-level images to more
accurately identify scenic and non-scenic locations.

%% file: beauty.bbl
\begin{thebibliography}{10}\itemsep=-1pt

\bibitem{arietta2014city}
S.~M. Arietta, A.~A. Efros, R.~Ramamoorthi, and M.~Agrawala.
\newblock City forensics: Using visual elements to predict non-visual city
  attributes.
\newblock {\em IEEE Transactions on Visualization and Computer Graphics},
  20(12):2624--2633, 2014.

\bibitem{avidan2007seam}
S.~Avidan and A.~Shamir.
\newblock Seam carving for content-aware image resizing.
\newblock {\em ACM Transactions on Graphics}, 26(3):10, 2007.

\bibitem{chandrasekaran2016we}
A.~Chandrasekaran, A.~K. Vijayakumar, S.~Antol, M.~Bansal, D.~Batra,
  C.~Lawrence~Zitnick, and D.~Parikh.
\newblock We are humor beings: Understanding and predicting visual humor.
\newblock In {\em IEEE Conference on Computer Vision and Pattern Recognition},
  2016.

\bibitem{deza2015understanding}
A.~Deza and D.~Parikh.
\newblock Understanding image virality.
\newblock In {\em IEEE Conference on Computer Vision and Pattern Recognition},
  2015.

\bibitem{icml2014c2_gardner14}
J.~Gardner, M.~Kusner, K.~Q. Weinberger, J.~Cunningham, and Z.~Xu.
\newblock Bayesian optimization with inequality constraints.
\newblock In {\em International Conference on Machine Learning}, 2014.

\bibitem{isola2011makes}
P.~Isola, J.~Xiao, A.~Torralba, and A.~Oliva.
\newblock What makes an image memorable?
\newblock In {\em IEEE Conference on Computer Vision and Pattern Recognition},
  2011.

\bibitem{jas2015image}
M.~Jas and D.~Parikh.
\newblock Image specificity.
\newblock In {\em IEEE Conference on Computer Vision and Pattern Recognition},
  2015.

\bibitem{jia2014caffe}
Y.~Jia, E.~Shelhamer, J.~Donahue, S.~Karayev, J.~Long, R.~Girshick,
  S.~Guadarrama, and T.~Darrell.
\newblock Caffe: Convolutional architecture for fast feature embedding.
\newblock In {\em ACM International Conference on Multimedia}, 2014.

\bibitem{karayev2014style}
S.~Karayev, M.~Trentacoste, H.~Han, A.~Agarwala, T.~Darrell, A.~Hertzmann, and
  H.~Winnemoeller.
\newblock Recognizing image style.
\newblock In {\em British Machine Vision Conference}, 2014.

\bibitem{ke2006design}
Y.~Ke, X.~Tang, and F.~Jing.
\newblock The design of high-level features for photo quality assessment.
\newblock In {\em IEEE Conference on Computer Vision and Pattern Recognition},
  2006.

\bibitem{laffont14transient}
P.-Y. Laffont, Z.~Ren, X.~Tao, C.~Qian, and J.~Hays.
\newblock Transient attributes for high-level understanding and editing of
  outdoor scenes.
\newblock {\em ACM Transactions on Graphics (SIGGRAPH)}, 33(4), 2014.

\bibitem{langlois2000maxims}
J.~H. Langlois, L.~Kalakanis, A.~J. Rubenstein, A.~Larson, M.~Hallam, and
  M.~Smoot.
\newblock Maxims or myths of beauty? a meta-analytic and theoretical review.
\newblock {\em Psychological bulletin}, 126(3):390, 2000.

\bibitem{lin2013cross}
T.-Y. Lin, S.~Belongie, and J.~Hays.
\newblock Cross-view image geolocalization.
\newblock In {\em IEEE Conference on Computer Vision and Pattern Recognition},
  2013.

\bibitem{lin2015learning}
T.-Y. Lin, Y.~Cui, S.~Belongie, and J.~Hays.
\newblock Learning deep representations for ground-to-aerial geolocalization.
\newblock In {\em IEEE Conference on Computer Vision and Pattern Recognition},
  2015.

\bibitem{lu2014rapid}
X.~Lu, Z.~Lin, H.~Jin, J.~Yang, and J.~Z. Wang.
\newblock Rapid: Rating pictorial aesthetics using deep learning.
\newblock In {\em ACM International Conference on Multimedia}, 2014.

\bibitem{lu2010photo2trip}
X.~Lu, C.~Wang, J.-M. Yang, Y.~Pang, and L.~Zhang.
\newblock Photo2trip: generating travel routes from geo-tagged photos for trip
  planning.
\newblock In {\em ACM International Conference on Multimedia}, 2010.

\bibitem{luo2008event}
J.~Luo, J.~Yu, D.~Joshi, and W.~Hao.
\newblock Event recognition: viewing the world with a third eye.
\newblock In {\em ACM International Conference on Multimedia}, 2008.

\bibitem{luo2008photo}
Y.~Luo and X.~Tang.
\newblock Photo and video quality evaluation: Focusing on the subject.
\newblock In {\em European Conference on Computer Vision}, 2008.

\bibitem{mcgranahan1999natural}
D.~A. McGranahan.
\newblock Natural amenities drive rural population change.
\newblock Technical report, United States Department of Agriculture, Economic
  Research Service, 1999.

\bibitem{patterson2012sun}
G.~Patterson and J.~Hays.
\newblock Sun attribute database: Discovering, annotating, and recognizing
  scene attributes.
\newblock In {\em IEEE Conference on Computer Vision and Pattern Recognition},
  2012.

\bibitem{lorenzo2015safety}
L.~Porzi, S.~Rota~Bul\`{o}, B.~Lepri, and E.~Ricci.
\newblock Predicting and understanding urban perception with convolutional
  neural networks.
\newblock In {\em ACM International Conference on Multimedia}, 2015.

\bibitem{quercia2015digital}
D.~Quercia, L.~M. Aiello, R.~Schifanella, and A.~Davies.
\newblock The digital life of walkable streets.
\newblock In {\em International World Wide Web Conference}, 2015.

\bibitem{quercia2014shortest}
D.~Quercia, R.~Schifanella, and L.~M. Aiello.
\newblock The shortest path to happiness: Recommending beautiful, quiet, and
  happy routes in the city.
\newblock In {\em ACM Conference on Hypertext and Social Media}, 2014.

\bibitem{runge2016no}
N.~Runge, P.~Samsonov, D.~Degraen, and J.~Sch{\"o}ning.
\newblock No more autobahn!: Scenic route generation using googles street view.
\newblock In {\em International Conference on Intelligent User Interfaces},
  2016.

\bibitem{seresinhe2017quantifying}
C.~I. Seresinhe, H.~S. Moat, and T.~Preis.
\newblock Quantifying scenic areas using crowdsourced data.
\newblock {\em Environment and Planning B: Urban Analytics and City Science},
  2017.

\bibitem{seresinhe2015quantifying}
C.~I. Seresinhe, T.~Preis, and H.~S. Moat.
\newblock Quantifying the impact of scenic environments on health.
\newblock {\em Scientific Reports}, 5, 2015.

\bibitem{seresinhe2017using}
C.~I. Seresinhe, T.~Preis, and H.~S. Moat.
\newblock Using deep learning to quantify the beauty of outdoor places.
\newblock {\em Royal Society Open Science}, 4(7), 2017.

\bibitem{su2011scenic}
H.-H. Su, T.-W. Chen, C.-C. Kao, W.~H. Hsu, and S.-Y. Chien.
\newblock Scenic photo quality assessment with bag of aesthetics-preserving
  features.
\newblock In {\em ACM International Conference on Multimedia}, 2011.

\bibitem{szegedy2015going}
C.~Szegedy, W.~Liu, Y.~Jia, P.~Sermanet, S.~Reed, D.~Anguelov, D.~Erhan,
  V.~Vanhoucke, and A.~Rabinovich.
\newblock Going deeper with convolutions.
\newblock In {\em IEEE Conference on Computer Vision and Pattern Recognition},
  2015.

\bibitem{van2009learning}
J.~Van De~Weijer, C.~Schmid, J.~Verbeek, and D.~Larlus.
\newblock Learning color names for real-world applications.
\newblock {\em IEEE Transactions on Image Processing}, 18(7):1512--1523, 2009.

\bibitem{planet}
T.~Weyand, I.~Kostrikov, and J.~Philbin.
\newblock Planet-photo geolocation with convolutional neural networks.
\newblock In {\em European Conference on Computer Vision}, 2016.

\bibitem{workman2015geocnn}
S.~Workman and N.~Jacobs.
\newblock On the location dependence of convolutional neural network features.
\newblock In {\em IEEE/ISPRS Workshop: Looking from above: When Earth
  observation meets vision}, 2015.

\bibitem{workman2015wide}
S.~Workman, R.~Souvenir, and N.~Jacobs.
\newblock Wide-area image geolocalization with aerial reference imagery.
\newblock In {\em IEEE International Conference on Computer Vision}, 2015.

\bibitem{workman2017unified}
S.~Workman, M.~Zhai, D.~J. Crandall, and N.~Jacobs.
\newblock A unified model for near and remote sensing.
\newblock In {\em IEEE International Conference on Computer Vision}, 2017.

\bibitem{workman2016horizon}
S.~Workman, M.~Zhai, and N.~Jacobs.
\newblock Horizon lines in the wild.
\newblock In {\em British Machine Vision Conference}, 2016.

\bibitem{xie2011im2map}
L.~Xie and S.~Newsam.
\newblock Im2map: deriving maps from georeferenced community contributed photo
  collections.
\newblock In {\em ACM SIGMM International Workshop on Social media}, 2011.

\bibitem{yeh2010personalized}
C.-H. Yeh, Y.-C. Ho, B.~A. Barsky, and M.~Ouhyoung.
\newblock Personalized photograph ranking and selection system.
\newblock In {\em ACM International Conference on Multimedia}, 2010.

\bibitem{zhai2017predicting}
M.~Zhai, Z.~Bessinger, S.~Workman, and N.~Jacobs.
\newblock Predicting ground-level scene layout from aerial imagery.
\newblock In {\em IEEE Conference on Computer Vision and Pattern Recognition},
  2017.

\bibitem{zhou2014object}
B.~Zhou, A.~Khosla, A.~Lapedriza, A.~Oliva, and A.~Torralba.
\newblock Object detectors emerge in deep scene cnns.
\newblock In {\em International Conference on Learning Representations}, 2014.

\bibitem{zhou2014places}
B.~Zhou, A.~Lapedriza, J.~Xiao, A.~Torralba, and A.~Oliva.
\newblock Learning deep features for scene recognition using places database.
\newblock In {\em Advances in Neural Information Processing Systems}, 2014.

\bibitem{zube1982landscape}
E.~H. Zube, J.~L. Sell, and J.~G. Taylor.
\newblock Landscape perception: research, application and theory.
\newblock {\em Landscape planning}, 9(1):1--33, 1982.

\end{thebibliography}
